\documentclass[10pt,twocolumn,letterpaper]{article}

\usepackage[pagenumbers]{cvpr} % To force page numbers, e.g. for an arXiv version

\usepackage{comment}
\definecolor{cvprblue}{rgb}{0.21,0.49,0.74}
\usepackage[pagebackref,breaklinks,colorlinks,allcolors=cvprblue]{hyperref}
\usepackage{algorithm}
\usepackage{algpseudocode}
\usepackage{indentfirst}
\usepackage{booktabs}
\usepackage{multirow}
\usepackage{array}
\usepackage{diagbox}
\usepackage{xcolor,colortbl}
\usepackage{graphicx}
\usepackage{makecell}
\usepackage{bbding}
\usepackage{float}
\usepackage{tablefootnote}
\usepackage{longtable}

\def\paperID{xxxxx}
\def\confName{CVPR}
\def\confYear{2026}

\title{MorphSeek: Fine-grained Latent Representation-Level Policy Optimization for Deformable Image Registration}

\author{Runxun Zhang$^{1,2,\dagger}$ \quad Yizhou Liu$^{1,3}$ \quad Dongrui Li$^4$ \quad Bo Xu$^{1,\ast}$ \quad Jingwei Wei$^{1,\dagger,\ast}$\\
$^1$Institute of Automation, Chinese Academy of Sciences, China\\
$^2$Sun Yat-sen University, China\\
$^3$Fudan University, China\\
$^4$Hebei Medical University, China\\
{\tt\small \{zhangrunxun2026, xubo, weijingwei2014\}@ia.ac.cn, liuyz25@m.fudan.edu.cn, llddrr@hebmu.edu.cn}\\
$^\dagger$Equal contribution \quad $^\ast$Corresponding author
}

\begin{document}
\maketitle
\begin{abstract}
\par Deformable image registration (DIR) remains a fundamental yet challenging problem in medical image analysis, largely due to the prohibitively high-dimensional deformation space of dense displacement fields and the scarcity of voxel-level supervision. Existing reinforcement learning frameworks often project this space into coarse, low-dimensional representations, limiting their ability to capture spatially variant deformations. We propose MorphSeek, a fine-grained representation-level policy optimization paradigm that reformulates DIR as a spatially continuous optimization process in the latent feature space. MorphSeek introduces a stochastic Gaussian policy head atop the encoder to model a distribution over latent features, facilitating efficient exploration and coarse-to-fine refinement. The framework integrates unsupervised warm-up with weakly supervised fine-tuning through Group Relative Policy Optimization, where multi-trajectory sampling stabilizes training and improves label efficiency. Across three 3D registration benchmarks (OASIS brain MRI, LiTS liver CT, and Abdomen MR–CT), MorphSeek achieves consistent Dice improvements over competitive baselines while maintaining high label efficiency with minimal parameter cost and low step-level latency overhead. Beyond optimizer specifics, MorphSeek advances a representation-level policy learning paradigm that achieves spatially coherent and data-efficient deformation optimization, offering a principled, backbone-agnostic, and optimizer-agnostic solution for scalable visual alignment in high-dimensional settings.
\end{abstract}

\section{Introduction}
\label{sec:intro}

Deformable image registration (DIR) is a highly challenging core task in medical image analysis\cite{0101,0102,0103}. Its goal is to establish voxel-wise spatial correspondences between two three-dimensional medical images, thereby enabling precise anatomical alignment. Owing to the pronounced non-rigid, large-scale deformations and inter-subject variability of anatomical structures, DIR is substantially more difficult than generic visual recognition tasks: it must achieve global structural alignment while preserving local geometric consistency\cite{0403zongshuDL2020}. Classical registration methods formulate DIR as a continuous optimization problem and solve for the deformation field via iterative procedures\cite{0201,0202,0203,0204},  but their computational cost is extremely high\cite{0301,0302}. 

Driven by the rapid progress of deep learning\cite{0401alexnet,0402attentionisallyouneed}, recent approaches adopt end-to-end encoder–decoder architectures to directly map image pairs to deformation fields, achieving significant gains in both efficiency and accuracy\cite{0501voxelmorph,0502transmorph,0503swinvoxelmorph,0504nicetrans,0505nicenet,0506pivit,0507transmatch}.

Nevertheless, deep learning--based DIR still faces two obstacles: \textbf{(i)} it relies heavily on supervision signals despite extremely limited annotations in most medical scenarios, and \textbf{(ii)} mainstream single-shot inference schemes remain challenged by complex large deformations, which ultimately limits registration accuracy.

The first challenge is to reliably solve high-difficulty, large-deformation registration problems given only a very small number of labeled examples. Complex anatomical structures and large-scale non-rigid deformations often require fine-grained voxel-level supervision to be stably aligned. However, segmentation annotations are exceedingly scarce in most medical settings. As a result, most registration models are forced to rely on unsupervised losses based on image similarity\cite{0701banSZ,0702ban,0703banFerrante,0704banlianhe}, whose ability to constrain local boundaries and subtle structures is limited. Existing works mainly strengthen unsupervised registration via pseudo-label generation\cite{0801SPACweibiaoqian,0806ontheflyweibiaoqian,0804weibiaoqianSynthMorph}, architectural refinements\cite{0802cascade,0803cascade,0805cyclemorph,0807diffVoxelmorph}, or new similarity metrics\cite{0808mindmetric,0809CNNmetric}, but comparatively little attention has been paid to maximizing supervision efficiency from a fixed yet very limited set of labels, especially for complex large-deformation cases.

The second challenge stems from the fact that most deep learning–based DIR models perform inference via a single forward pass, i.e., they predict the deformation field in one shot\cite{0501voxelmorph,0502transmorph,0807diffVoxelmorph}. In scenarios involving large-scale non-rigid deformations, such as thoracic or abdominal registration, such models often fit only the global structural differences while struggling to reliably recover local boundaries and fine geometric details. To improve alignment under complex deformations, several methods introduce step-wise registration\cite{0901VRNetfine,0902fine,0903fine,0904RIIRstep,0905LapIRN}, decomposing a large deformation into a sequence of incremental updates to realize coarse-to-fine optimization. However, existing step-wise frameworks typically rely on manually designed, fixed cascaded structures and lack a learnable multi-step decision policy. Reinforcement learning (RL) has been explored for image registration because its stochastic, Markov decision process is naturally compatible with step-wise optimization. However, most existing RL-based registration methods are confined to low-dimensional rigid transformations\cite{1101RL_G,1102RL_G,1103RL_G,1104RL_G,1105RL_G,1106RL_G}. Directly treating a full 3D deformable field as the action space would make memory consumption and sampling cost prohibitive, severely limiting the applicability of RL to real-world deformable registration.

To address these issues, we propose MorphSeek, which reformulates deformable registration as latent-space policy optimization by introducing a sampleable high-resolution latent representation at the top encoder layer and treating it as the policy action, thereby avoiding RL reasoning directly in the million-dimensional deformation field while preserving fine spatial granularity. The framework first performs unsupervised warm-up to shape a stable latent space and then applies Group Relative Policy Optimization (GRPO) with multiple trajectories and multiple steps under weak supervision, repeatedly reusing scarce labels for coarse-to-fine refinement. To make such high-dimensional policies trainable, we further propose Latent-Dimension Variance Normalization (LDVN), which controls the variance of log-likelihoods and provides direction-preserving, scale-controlled policy updates for scalable 3D dense prediction.

Our main contributions are as follows:

\begin{itemize}
	\item We introduce a new latent-space policy optimization paradigm for deformable image registration. By defining the policy distribution in the encoder latent feature space instead of operating directly on the dense deformation field, we realize a fine-grained, scalable, and backbone-agnostic step-wise optimization mechanism.
	\item We propose LDVN, a statistical normalization scheme that stabilizes GRPO in high-dimensional dense prediction settings. We show that LDVN controls the variance of the log-likelihood while preserving policy-gradient direction, allowing GRPO to operate stably in high-dimensional 3D feature spaces and providing both theoretical and practical support for applying RL to dense prediction tasks.
	\item We construct a highly label-efficient multi-trajectory, multi-step weakly supervised framework. Through warm-up pre-training and GRPO-guided coarse-to-fine refinement, our framework repeatedly reuses supervision signals under very limited annotations, markedly improving large-deformation registration quality while maintaining a comparable parameter count and acceptable inference latency.
\end{itemize}

\section{Related Work}
\label{sec:rela}

\subsection{DL-Based Deformable Medical Image Registration}

\par Early DL-based DIR methods were fully supervised using deformation vector fields (DVFs)\cite{0601quanRGS,0602quanCTS,0603quanCTS,0604quanRGS_RL,0605quanRGS,0606quanCTS}. After Hu et al. proposed using anatomical segmentations instead of DVFs for supervision\cite{0701banSZ}, such fully supervised strategies became less common. Since VoxelMorph\cite{0501voxelmorph}, a U-Net-style CNN trainable in an unsupervised manner, subsequent studies have largely evolved within this unsupervised U-Net-style paradigm\cite{0504nicetrans,0807diffVoxelmorph,0505nicenet,0507transmatch,0502transmorph,0503swinvoxelmorph}. 

Meanwhile, leveraging segmentation labels to further improve registration has become an active research direction. Hu et al. extended the label-driven idea and systematically discussed the advantages of segmentation-supervised training over purely unsupervised objectives\cite{0702ban}. Ferrante et al. used segmentation labels to guide the weighting of different similarity terms during registration\cite{0703banFerrante}. Zhou et al. proposed macJNet\cite{0704banlianhe}, which jointly learns two segmentation networks and one registration network.

There have also been attempts to combine unsupervised and weakly-supervised learning. Li et al. combined segmentation-labeled and unlabeled image pairs for registration using consistency regularization in a student-teacher framework\cite{1201semi}. Unsupervised models such as VoxelMorph\cite{0501voxelmorph} often include hybrid objectives that combine image-similarity and label-based losses. Chen et al. proposed a training strategy that first performs unsupervised pretraining with randomly generated images and then fine-tunes on the target task, maintaining strong performance when domain-specific data are limited\cite{1202}. However, these approaches still do not simultaneously exploit both unlabeled and labeled data from the same domain to maximally optimize the registration model.

From another perspective, coarse-to-fine registration has been extensively explored. ULAE-Net\cite{1301ulaenet} performs step-wise registration by repeatedly applying the network. LapIRN\cite{0905LapIRN} cascades multiple Laplacian pyramid networks to implement coarse-to-fine alignment. However, these methods use fixed, deterministic schedules and lack adaptive exploration of optimal registration strategies.

\subsection{Reinforcement Learning in DIR}

\par In recent years, reinforcement learning (RL) has advanced rapidly in decision-making, robotics, and large-model reasoning\cite{1002,1001deepseek}. However, when applied to dense prediction tasks, the continuous and high-dimensional state and action spaces—particularly in 3D—lead to unstable training, high exploration cost, and substantial memory overhead; this bottleneck is especially pronounced in deformable image registration (DIR).

Krebs et al. proposed an agent-based non-rigid registration framework that reduces the DIR action space from dense DVFs to a statistical deformation model (PCA over B-spline–parameterized deformations), significantly lowering the action dimensionality\cite{0604quanRGS_RL}. However, their method requires dense DVFs as supervision, which is impractical in contemporary settings. Luo et al. introduced SPAC\cite{0801SPACweibiaoqian}, which compresses a pair into a 64-D plan and relies on an extra critic for stability (SAC-based), which complicates deployment. Moreover, the 64-D bottleneck discards spatial detail, limiting performance.

In summary, the central challenge for applying RL to DIR is how to retain model generality while effectively shifting exploration and optimization from the dense field to a low-dimensional, training-friendly space.

\begin{comment}
    第一句：强化学习的概念。在Dense Prediction上应用的难点。

    Krebs et al

    SPAC

    分割PO
\end{comment}

\begin{comment}
早期的DL-based配准方法使用完全监督的方式，需要密集形变场作为GT标签，它们通常由传统的配准方法生成。在Hu等人提出使用分割代替形变场作为标签后，完全监督的方式被逐渐抛弃。2018年，Balakrishnan等人提出了著名的VoxelMorph，一种基于CNN的网络U-Net型架构。这种架构可以仅使用图像相似度和形变正则化损失的无监督方式进行训练。后续的研究不断在“无监督+U-Net Style”框架中演进。Chen等人提出了TransMorph，在编码器中引入Swin-Transformer网络；Zhu等人提出的Swin-VoxelMorph则完全使用Swin-Transformer来构建编码器和解码器。

与此同时，如何更好地利用分割标签改进配准也成为研究热点。Hu等人扩展了分割标签驱动的思路，系统讨论了分割标签监督相比无监督的优势。E. Ferrante等人利用分割标签指导配准时各个相似性度量权重的选择。Zhou等人提出了macJNet，将两个分割网络和一个配准网络组合进行联合学习。

在无监督和半监督的结合上，我们也发现了前人的尝试。VoxelMorph等无监督模型通常提供了image-similarity + label-based loss的混合损失接口；Chen等人提出了使用随机生成的图像进行预训练，再在目标任务上微调的训练方法，在领域数据有限的情况下保持高性能。然而，这些尝试都未能同时利用领域内的无标签和标签数据，对配准模型进行最大化的优化。

在另一个角度，coarse-to-fine的配准模型也不断被探索(cascade)。Shu等人提出了ULAE-net，通过多次运行网络的办法step-wise地进行配准。Mok等人提出了LapIRN，通过将多个laplacian pyramid networks级联从而进行coarse-to-fine的配准。
\end{comment}
% \input{sec/2_formatting}
\section{Method}
\label{sec:me}

\begin{figure*}[!ht]
  \centering
  %\fbox{\rule{0pt}{3in} \rule{0.8\linewidth}{0pt}}
   \includegraphics[width=0.8\linewidth]{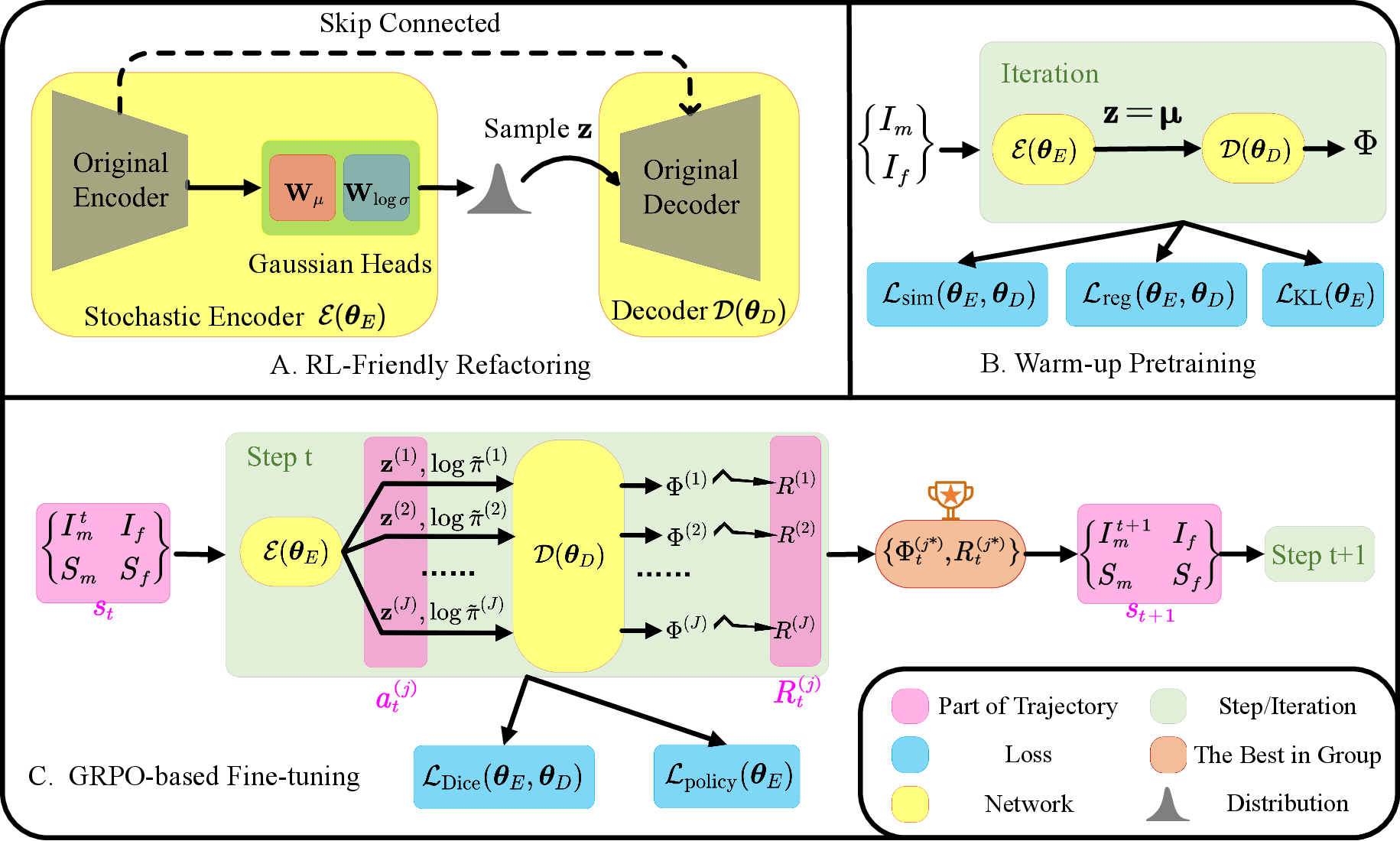}
   \caption{MorphSeek Registration Framework Process}
   \label{02}
\end{figure*}

MorphSeek is a training paradigm that can be generalized to any encoder–decoder–based registration model and provides a unified formulation of deformable registration via latent-space policy optimization. The paradigm consists of three stages: \textbf{(i)} an RL-friendly refactoring step that constructs a sampleable latent space at the top of the encoder and decouples the encoder and decoder, \textbf{(ii)} an unsupervised warm-up phase that shapes a stable latent-space structure, and \textbf{(iii)} a GRPO-based weakly supervised fine-tuning phase that performs coarse-to-fine policy updates with multiple trajectories and multiple steps so that the scarce labels can be repeatedly reused (Figure~\ref{02}). For clarity of exposition, we instantiate MorphSeek with a U-Net backbone.

\subsection{Refactoring Registration Networks for Latent Policy Learning}

\par Deformable registration networks typically adopt a U-Net-style encoder-decoder architecture with skip connections. Given a moving image $I_m$ and fixed image $I_f$ both in $\mathbb{R}^{H \times W \times D}$, the network takes their concatenation $x = [I_m, I_f]$ as input. The encoder $\mathcal{E}$ extracts multi-scale features:

\begin{equation}  
    \{\mathbf{f}_1, \mathbf{f}_2, \dots, \mathbf{f}_L\} = \mathcal{E}(x)  
\end{equation}  

where $\mathbf{f}_l \in \mathbb{R}^{C_l \times H_l \times W_l \times D_l}$ denotes the feature at level $l$ with progressively reduced spatial resolution. The decoder $\mathcal{D}$ then upsamples and fuses these features via skip connections to predict a dense deformation field:

\begin{equation}  
    \Phi = \mathcal{D}(\{\mathbf{f}_1, \mathbf{f}_2, \dots, \mathbf{f}_L\}) \in \mathbb{R}^{3 \times H \times W \times D}
    \label{decoder}  
\end{equation}  

where $\Phi$ represents per-voxel displacement vectors, yielding the warped image $I_m \circ \Phi$.

\par In a conventional deterministic encoder $\mathcal{E}$, the top-level feature $\mathbf{f_L}$ is directly fed into the decoder $\mathcal{D}$. To enable RL-based fine-tuning, we decouple the encoder and decoder in the U-Net-style architecture (while preserving skip connections) and introduce stochasticity at the encoder output via Gaussian parameterization. This allows the encoder to model a probability distribution over latent vectors and supports policy optimization with Group Relative Policy Optimization (GRPO).  

Specifically, we append two convolutional heads to the top-level feature $\mathbf{f_L} \in \mathbb{R}^{C_L \times H_L \times W_L \times D_L}$: a mean head $\mathbf{W}_{\mu}$ and a log-standard-deviation head $\mathbf{W}_{\log\sigma}$, both with kernel size 1, i.e., $\mathbf{W}_{\mu}, \mathbf{W}_{\log\sigma} \in \mathbb{R}^{C_L \times C_L \times 1 \times 1 \times 1}$.  

These two heads take the tensor $\mathbf{f_L}$ and parameterize a multivariate Gaussian $\mathcal{N}(\boldsymbol{\mu}, \boldsymbol{\sigma^2})$. To stabilize training, we impose constraints and clipping on the outputs:  

\begin{align}
\boldsymbol{\mu} &= \tanh(\mathbf{W}_\mu(\mathbf{f_L})) \cdot \lambda_{\text{scale}} 
    \in \mathbb{R}^{C_L \times H_L \times W_L \times D_L} \label{mu}\\
\log \boldsymbol{\sigma} &= \text{clip}(\mathbf{W}_{\log\sigma}(\mathbf{f_L}), \sigma_{min}, \sigma_{max})
    \in \mathbb{R}^{C_L \times H_L \times W_L \times D_L}\label{logsigma}
\end{align}

We further introduce a temperature parameter $\tau > 0$ to modulate exploration. During training, the latent vector is sampled using the reparameterization trick:  

\begin{equation}  
   \mathbf{z} = \boldsymbol{\mu} + \tau \cdot \boldsymbol{\sigma} \odot \boldsymbol{\epsilon}, \quad \boldsymbol{\epsilon} \sim \mathcal{N}(\mathbf{0}, \mathbf{I})  
   \label{refrac}
\end{equation}  

Compared to Eq.~\ref{decoder}, the input–output of the decoder is modified as:  
\begin{equation}  
    \Phi = \mathcal{D}(\{\mathbf{f_1}, \mathbf{f_2}, \dots, \mathbf{f_{L-1}}, \mathbf{z}\})  
\end{equation}

\subsection{Warm-up Priors for Stable Policy Optimization} 
\label{warmuppriors}

\par To ensure that the subsequent GRPO fine-tuning operates in a stable and well-conditioned latent space, we first pretrain the encoder and decoder on unlabeled data. Comparative analyses with and without warm-up are reported in Section~\ref{sec:ra}.  

During warm-up, to obtain stable warping estimates, we adopt a deterministic latent variable by setting $\tau = 0$, i.e.,  
\begin{equation}  
\mathbf{z} = \boldsymbol{\mu}.  
\label{z}
\end{equation}  
This deterministic warm-up forces anatomical information into the mean code before policy sampling and empirically reduces posterior-collapse risk, preserving the stochastic output variance later required by GRPO exploration.

Let $\boldsymbol{\theta}_E$ and $\boldsymbol{\theta}_D$ denote the trainable parameters of the encoder and decoder, respectively, and let $\boldsymbol{\theta}=\{\boldsymbol{\theta}_E,\boldsymbol{\theta}_D\}$. The overall warm-up objective minimizes an unsupervised loss composed of an image-similarity term, a deformation regularizer, and a KL penalty on the Gaussian heads:  
\begin{equation}  
\begin{split}
\mathcal{L}_{\text{warm}}(\boldsymbol{\theta})
= &\; \mathcal{L}_{\text{sim}}(I_f, I_{m \circ \Phi})
+ \lambda_{\text{reg}} \, \mathcal{L}_{\text{reg}}(\Phi) \\
&+ \beta_{\text{KL}} \, \mathcal{L}_{\text{KL}}\big(
q_{\boldsymbol{\theta}_E}(\mathbf{z} \mid \mathbf{f}_L)
\,\|\, \mathcal{N}(\mathbf{0},\mathbf{I})\big),
\end{split}
\label{warmup}  
\end{equation}  
where $\lambda_{\text{reg}}$ and $\beta_{\text{KL}}$ are weighting coefficients, and $q_{\boldsymbol{\theta}_E}$ denotes the factorized Gaussian parameterized by the encoder. Each loss component can be instantiated in multiple ways; for this work, we use mean squared error (MSE), diffusion regularization, and standard KL divergence, with exact formulations detailed in the supplement.

\subsection{Multi-Trajectory GRPO for Step-Wise Registration}  
\label{subsec:grpo}  

\par After warm-up pretraining, we fine-tune the encoder–decoder with segmentation labels under the GRPO framework to further improve registration accuracy. In this stage, the encoder's stochastic output distribution is treated as a latent policy, parameterized as $\pi(\mathbf{z}\mid\boldsymbol{\mu},\boldsymbol{\sigma})$ from the current state feature $\mathbf{f}_L$, where the state $s_t$ is the current registration pair $\{I_m^{t-1}, I_f\}$ and the action $a_t$ is the sampled latent $\mathbf{z}$. Here, $t$ denotes the refinement step within one forward pass, and we initialize the cumulative deformation as $\Phi_0=\mathrm{Id}$. At each fine-tuning step $t$, we generate a group of trajectories per sample to enable exploration through encoder stochasticity.

For each trajectory $j=1,\dots,J$, the decoder produces a single-step deformation field $\phi^{(j)}_t$ from the sampled latent $\mathbf{z}^{(j)}$, and we compute a scalar reward:  
\begin{equation}  
\begin{split}
R^{(j)} \;=\; & w_{\text{Dice}} \cdot [\mathrm{Dice}\!\big(S_f,\, S_{m \circ \Phi^{(j)}_t}\big)-\mathrm{Dice}\!\big(S_f,\, S_{m \circ \Phi_{t-1}}\big)] \\ & \;+\; w_{\text{NJD}} \cdot \text{NJD}\!\big(\Phi^{(j)}_t\big),  
\end{split}
\end{equation}  

where $S_f, S_m \in \mathbb{R}^{K \times H \times W \times D}$ are the fixed and moving segmentation labels, and $\Phi^{(j)}_t=\Phi_{t-1}\circ\phi^{(j)}_t$. Here, $\text{NJD}$ penalizes voxels with negative Jacobian determinants $|J_{\Phi^{(j)}_t}|<0$, and $w_{\text{Dice}}>0$, $w_{\text{NJD}}<0$ are scalar weights.  

To compute policy gradients, we perform group-wise normalization of the trajectory rewards for each sample, yielding the advantage  
\begin{equation}  
A^{(j)} \;=\; \frac{R^{(j)} - \bar{R}}{\sigma_R + \epsilon},  
\label{A_j}
\end{equation}  
where $\bar{R}$ and $\sigma_R$ are the mean and standard deviation of $\{R^{(j)}\}_{j=1}^J$ for the current sample, and $\epsilon = 10^{-8}$ prevents division by zero. This per-sample normalization also acts as an implicit hard-case reweighting: without it, easy cases with large absolute gains would dominate the gradients, whereas standardized advantages preserve learning signal from anatomically difficult pairs. We also compute the relative log-likelihood within the group:  
\begin{equation}  
\log \tilde{\pi}^{(j)} \;=\; \log \pi\!\big(\mathbf{z}^{(j)} \,\big|\, \boldsymbol{\mu}, \boldsymbol{\sigma}\big) \;-\; \overline{\log \pi}, 
\label{logpimean}
\end{equation}  
where $\boldsymbol{\mu}, \boldsymbol{\sigma}$ are the policy parameters produced from the current state feature $\mathbf{f}_L$ (shared across the $J$ samples), and $\overline{\log \pi}$ is the group mean of $\log \pi(\mathbf{z}^{(j)} \mid \boldsymbol{\mu}, \boldsymbol{\sigma})$.  

However, for conventional backbones\cite{0502transmorph,0504nicetrans}, the latent dimensionality $N=C_L\times H_L\times W_L\times D_L$ often reaches tens of thousands, which is far larger than in typical GRPO applications. Directly summing over all latent dimensions can make within-group relative log-likelihoods numerically unstable, weakening exploration discrimination and destabilizing training.

We therefore introduce Latent-Dimension Variance Normalization (LDVN), which rescales the log-likelihood as
\begin{equation}
\begin{split}
    \log \pi&(\mathbf{z} \mid \boldsymbol{\mu}, \boldsymbol{\sigma}) \\
    &= -\frac{1}{2s} \sum_{i=1}^{N} \left[
        \left( \frac{z_i - \mu_i}{\tau \sigma_i} \right)^2
        + \log (2\pi \tau^2 \sigma_i^2)
      \right].
\end{split}
\label{logpi}
\end{equation}
where $s$ controls the scale of within-group relative log-likelihoods across different latent dimensionalities. In practice, setting $s\propto\sqrt{N}$ keeps GRPO updates numerically stable while preserving within-group ordering and the policy-gradient direction. LDVN only affects the policy-loss statistics and does not alter the sampling distribution of $\pi(\mathbf{z}\mid\boldsymbol{\mu},\boldsymbol{\sigma})$ or the exploration temperature $\tau$. Detailed variance analysis and empirical verification are provided in the supplement.

\begin{algorithm}[t]
\caption{GRPO-based Fine-tuning}
\label{grpo}
\begin{algorithmic}
\State \textbf{Initialize:} Load pretrained $\boldsymbol{\theta} = (\boldsymbol{\theta_E}, \boldsymbol{\theta_D})$, set $\tau > 0$
\For{each pair $\{I_m, I_f, S_m, S_f\}$}
    \State \textbf{Initialize:} $I_m^{0} \gets I_m$
    \For{each step $t = 1, \dots, T$}
        \State \textbf{Sample} $\phi^{(1)}_t, \phi^{(2)}_t, \dots, \phi^{(J)}_t$
        \State \textbf{Compute} $\Phi^{(j)}_t = \Phi_{t-1} \circ \phi^{(j)}_t$
        \State \textbf{Compute} $R^{(j)}$, $A^{(j)}$, $\log\tilde{\pi}^{(j)}$, and $\mathcal{L}_{\text{grpo}}$
        \State \textbf{Update} $\boldsymbol{\theta} \gets \boldsymbol{\theta} - \eta\nabla_{\boldsymbol{\theta}}\mathcal{L}_{\text{grpo}}$
        \State \textbf{Update} $\Phi_{t}, I_m^{t} $ via Eqs.~\ref{update_fai} and \ref{update_I}
    \EndFor
\EndFor
\end{algorithmic}
\end{algorithm}

In practice, we set $s=\sqrt{N}$ by default and defer the variance derivation, weak-dependence analysis, and ablations to the supplement.

In all, the policy loss is defined as  
\begin{equation}  
\mathcal{L}_{\text{policy}}(\boldsymbol{\theta}_E) \;=\; -\,\frac{1}{J}\,\sum_{j=1}^{J} \; A^{(j)} \cdot \log \tilde{\pi}^{(j)}, 
 \label{losspolicy}
\end{equation}  
which updates the encoder parameters $\boldsymbol{\theta}_E$ through the gradient of $\log \pi$, increasing the sampling probability of high-reward trajectories.  

In parallel, we compute a supervised soft-Dice loss using differentiable warping:  
\begin{equation}  
\mathcal{L}_{\text{Dice}}(\boldsymbol{\theta}) \;=\; \frac{1}{J}\,\sum_{j=1}^{J}\,\Big[\,1 - \mathrm{Dice}\!\big(S_f,\, S_{m \circ \Phi^{(j)}_t}\big)\Big]. 
\label{tra_Dice}
\end{equation}  
Note that $\mathcal{L}_{\text{Dice}}$ is computed with soft labels via trilinear interpolation to ensure differentiability, whereas the reward in $\mathcal{L}_{\text{policy}}$ does not backpropagate gradients and is computed with hard labels to more faithfully reflect the task metric.  
While $\mathcal{L}_{\text{Dice}}$ supplies one deterministic overlap signal, $\mathcal{L}_{\text{policy}}$ ranks $J$ sampled hypotheses at each of $T$ refinement steps for the same labeled pair. Each pair therefore yields $T\times J$ relative supervision events, which helps explain MorphSeek's label-efficiency gains.

To prevent catastrophic forgetting of the representations learned during warm-up and to maintain smooth and physically plausible deformations, we retain the warm-up objective as a regularizer during GRPO. Beyond stabilization, $\mathcal{L}_{\text{warm}}$ acts as an anatomy-preserving prior that keeps optimization close to the warm-up manifold, reducing reward hacking and implausible but numerically favorable deformations. It is computed through Eqs.~\ref{z} and \ref{warmup} rather than sample averaging. The overall loss for GRPO fine-tuning is\footnote{The encoder parameterizes a Gaussian distribution $\mathcal{N}(\boldsymbol{\mu}, \boldsymbol{\sigma}^2)$ in both stages. During warm-up, we regularize it toward $\mathcal{N}(\mathbf{0}, \mathbf{I})$ via the KL term in Eq.~\ref{warmup}. In GRPO fine-tuning, this same KL divergence (evaluated with $\tau=0$ as in warm-up) is retained to maintain a fixed-prior trust region.}    
\begin{equation}  
\mathcal{L}_{\text{grpo}}(\boldsymbol{\theta}) \;=\; \mathcal{L}_{\text{policy}}(\boldsymbol{\theta}_E) \;+\; \lambda_{\text{warm}}\,\mathcal{L}_{\text{warm}}(\boldsymbol{\theta}) \;+\; \lambda_{\text{Dice}}\,\mathcal{L}_{\text{Dice}}(\boldsymbol{\theta}).  
\end{equation}

Unlike PPO/TRPO which bound consecutive policy ratios, 
we adopt a fixed-prior trust region by penalizing 
$\mathrm{KL}(\pi_{\boldsymbol{\theta}_E}\,\|\,\mathcal N(0,\mathbf{I}))$ with a target-KL schedule. Warm-up already puts $\pi_{\boldsymbol{\theta}_{E0}}$ near $\mathcal N(0,\mathbf{I})$, 
so keeping this KL small bounds the drift to the warm-up policy 
while remaining critic-free and ratio-free in high-dimensional latents.

At the end of each step, we greedily select the trajectory with the highest reward to update the current state. Specifically, letting $j^* = \arg\max_{j} R^{(j)}$, we update the deformation field and moving image via  

\begin{equation}  
\Phi_{t} \;\gets\; \Phi_{t-1}\circ \phi^{(j^*)},
\label{update_fai}
\end{equation}

\begin{equation}  
I_m^{t} \;\gets\; I_{m \circ \Phi_t}.
\label{update_I}
\end{equation}  
The process is repeated for $T$ steps or until convergence. The overall procedure is summarized in Algorithm~\ref{grpo}.

\section{Experiments}
\label{sec:exp}

\begin{table*}
\centering
\caption{Quantitative comparison on three registration tasks. All methods except affine \textbf{use weakly supervised training}. $\uparrow$: higher is better; $\downarrow$: lower is better. Our results are shown in bold and marked with * if there is a statistically significant difference (p \textless~0.05) from their baselines by a Wilcoxon signed-rank test. In MorphSeek, both Trajs/Steps are set to 6/3. NJDs in SPAC cannot be calculated due to coupling in the deformation field; see appendix.}
\label{tab:main_results}
\scalebox{0.82}{
\begin{tabular}{l|cc|cc|cc}
\toprule
\multirow{2}{*}{Method} & \multicolumn{2}{c|}{OASIS (Brain MRI)} & \multicolumn{2}{c|}{LiTS (Liver CT)} & \multicolumn{2}{c}{Abdomen MR$\leftarrow$CT} \\
 & Mean Dice (\%) $\uparrow$ & NJD (\%) $\downarrow$ & Dice (\%) $\uparrow$ & NJD (\%) $\downarrow$ & Mean Dice (\%) $\uparrow$ & NJD (\%) $\downarrow$ \\
\midrule
Only Affine & 58.5$\pm$4.0 & -- & 60.2$\pm$10.0 & -- & 37.8$\pm$18.1 & -- \\
\midrule
CorrMLP~\cite{1501corrMLP} & 88.3$\pm$1.3 & 0.0$\pm$0.0 & 89.2$\pm$3.0 & 0.2$\pm$0.1 & 86.8$\pm$5.0 & 0.4$\pm$0.4 \\
RIIR~\cite{0904RIIRstep} (Steps=12) & 87.7$\pm$2.5 & 0.1$\pm$0.0 & 88.9$\pm$4.1 & 0.3$\pm$0.1 & 80.7$\pm$4.3 & 1.0$\pm$0.8 \\
WarpDDF+RegCut~\cite{1201semi} & 86.6$\pm$3.8 & 0.2$\pm$0.0 & 85.5$\pm$3.9 & 0.6$\pm$0.1 & 85.4$\pm$8.2 & 1.1$\pm$0.5 \\
SPAC~\cite{0801SPACweibiaoqian} (Steps=20) & 78.9$\pm$5.3 & N/A & 75.3$\pm$8.3 & N/A & 69.2$\pm$10.1 & N/A \\
\midrule
VoxelMorph-L~\cite{0501voxelmorph} & 84.7$\pm$2.4 & 0.1$\pm$0.1 & 84.9$\pm$6.3 & 0.7$\pm$0.1 & 77.9$\pm$9.1 & 1.0$\pm$0.6 \\
\quad + MorphSeek (Ours) & \textbf{87.2$\pm$2.0}* & \textbf{0.1$\pm$0.0}* & \textbf{89.0$\pm$3.1}* & \textbf{0.2$\pm$0.1}* & \textbf{82.4$\pm$6.4}* & \textbf{0.6$\pm$0.4}* \\
TransMorph~\cite{0502transmorph} & 85.8$\pm$1.4 & 0.1$\pm$0.0 & 88.3$\pm$5.3 & 0.4$\pm$0.1 & 82.3$\pm$4.8 & 0.8$\pm$0.4 \\
\quad + MorphSeek (Ours) & \textbf{88.9$\pm$1.8}* & \textbf{0.1$\pm$0.0}* & \textbf{90.1$\pm$4.8}* & \textbf{0.2$\pm$0.1}* & \textbf{86.5$\pm$3.4}* & \textbf{0.4$\pm$0.2}* \\
NICE-Trans~\cite{0504nicetrans} & 86.7$\pm$2.3 & 0.0$\pm$0.0 & 88.4$\pm$3.9 & 0.1$\pm$0.0 & 83.1$\pm$3.8 & 0.3$\pm$0.1 \\
\quad + MorphSeek (Ours) & \textbf{89.0$\pm$1.5}* & \textbf{0.0$\pm$0.0} & \textbf{90.5$\pm$3.7}* & \textbf{0.2$\pm$0.1} & \textbf{86.5$\pm$3.0}* & \textbf{0.3$\pm$0.2}* \\
\bottomrule
\end{tabular}}
\end{table*}

\subsection{Datasets}
\par We evaluate MorphSeek on three 3D registration tasks: OASIS brain MRI, LiTS liver CT, and Abdomen MR$\leftarrow$CT. We first split volumes/scans into disjoint train/validation/test pools and then construct non-overlapping pair lists only within each pool, avoiding test-volume leakage. For OASIS / LiTS / Abdomen MR$\leftarrow$CT, we use 400/100/20 pretraining/GRPO/validation pairs, with 19 official validation pairs, 40 held-out test pairs, and 8 official paired scans for testing, respectively. Exact preprocessing, resampling, and pair-construction details are provided in the supplement. For the cross-modality task, we replace MSE with the MIND descriptor\cite{0808mindmetric}.

\subsection{Baselines and Implementations}
\par We refactor VoxelMorph-L, TransMorph, and NICE-Trans under the same weakly supervised setting, using identical pair lists, the same labeled pairs, and comparable epoch budgets. We further compare against CorrMLP\cite{1501corrMLP}, RIIR\cite{0904RIIRstep}, SPAC\cite{0801SPACweibiaoqian}, and WarpDDF+RegCut\cite{1201semi}; for methods whose public code or training recipes are not directly compatible with our unified setting, we follow their published protocols. VoxelMorph-L uses channels [32, 64, 128, 256, 256] to provide a sufficiently expressive latent space for GRPO. Training uses Adam (1e-4) with batch size 1; detailed hyperparameters and hardware are given in the supplement.

We report Dice\cite{1701dice} (\%) for segmentation overlap and the percentage of voxels with negative Jacobian determinant (NJD\cite{1703njd}, \%) for deformation regularity.

\section{Results and Analysis}
\label{sec:ra}

\begin{figure*}[t]
  \centering
   \includegraphics[width=0.96\linewidth]{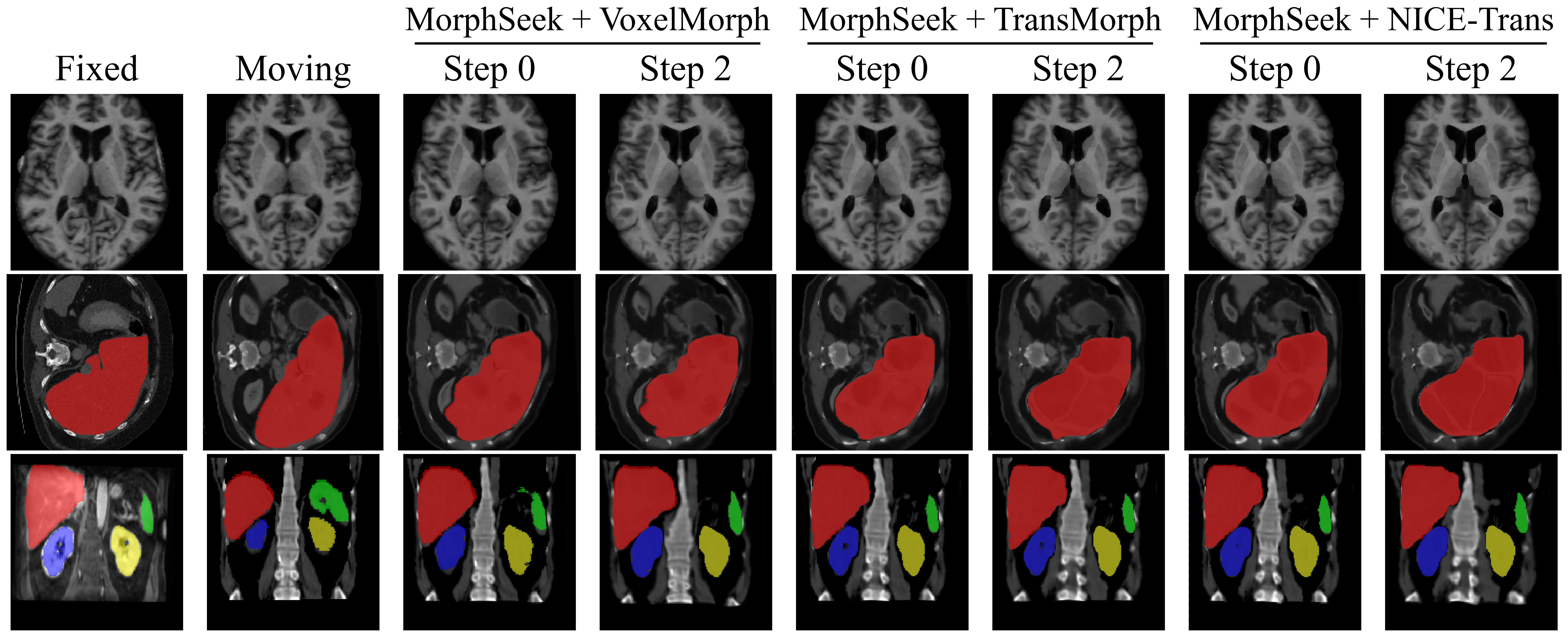}
   \caption{Representative visual comparisons across the three registration tasks. Labels are overlaid only for the two abdominal tasks; OASIS is left unlabeled to avoid clutter from its 35 foreground classes. Additional visual results are provided in the supplement.}
   \label{03}
\end{figure*}

\subsection{Overall Performance Across Tasks and Backbones}

As summarized in Fig.~\ref{03} and Table~\ref{tab:main_results}, MorphSeek consistently improves Dice and reduces NJD across three 3D benchmarks and three backbones (VoxelMorph-L, TransMorph, and NICE-Trans). On OASIS, Dice increases by 2–3\% while NJD decreases by roughly one-third relative to the corresponding baselines; on the more challenging cross-modality Abdomen MR$\leftarrow$CT task, MorphSeek yields more than a 4\% Dice gain and nearly halves NJD for TransMorph. Most gains are statistically significant under the Wilcoxon signed-rank test (p \textless~0.05), indicating that latent-space policy optimization benefits both small-deformation brain MRI registration and large-deformation cross-modality scenarios in terms of global alignment and local regularity. Compared with other multi-stage alternatives, this advantage is not limited to one backbone: MorphSeek also outperforms RIIR in Table~\ref{tab:main_results}, and supplementary results on OASIS show that it remains stronger than LapIRN-stage3 under the same 100-pair labeled setting.

\subsection{Policy \& Label Efficiency Ablation}

The ablation on trajectory number and refinement steps on OASIS (Table~\ref{tab:trajs_steps_oasis}) reveals a clear pattern. Increasing the number of trajectories up to six yields a steady improvement in Dice and a decrease in NJD, while adding refinement steps from one to three also brings consistent gains. Beyond three steps, however, the benefits saturate and the deformation field starts to show artifacts, reflected by degraded NJD and local distortions.

\begin{table}[t]
\centering
\caption{Ablation study on trajectory number and refinement steps on OASIS dataset. Using TransMorph + MorphSeek. Each cell shows Dice (\%) $\uparrow$ / NJD (\%) $\downarrow$.}
\label{tab:trajs_steps_oasis}
\scalebox{0.75}{
\begin{tabular}{c|cccc}
\toprule
\diagbox{\textbf{Trajs}}{\textbf{Steps}} & \textbf{1} & \textbf{2} & \textbf{3} & \textbf{4} \\
\midrule
\textbf{2} & 86.71 / 0.08 & 87.13 / 0.08 & 87.78 / 0.08 & 87.94 / 0.08 \\
\textbf{4} & 86.89 / 0.07 & 87.96 / 0.06 & 88.26 / 0.06 & 88.14 / 0.08 \\
\textbf{6} & 87.67 / 0.06 & 88.72 / 0.05 & \textbf{88.89 / 0.06} & 88.51 / 0.07 \\
\textbf{8} & OOM Error & N/A & N/A & N/A \\
\bottomrule
\end{tabular}}
\end{table}

This behavior is consistent with the intended coarse-to-fine design: the first step focuses on establishing coarse alignment, whereas subsequent steps repeatedly enforce local constraints under the same labels, effectively reusing weak supervision. When the data have already been “fully exploited,” additional steps no longer help and instead tend to compromise the physical plausibility of the deformation. Moreover, attempting more than 8 trajectories leads to out-of-memory errors, which matches the increased sampling cost in a high-dimensional policy space.

\begin{figure}[t]
  \centering
  %\fbox{\rule{0pt}{2in} \rule{0.8\linewidth}{0pt}}
   \includegraphics[width=0.75\linewidth]{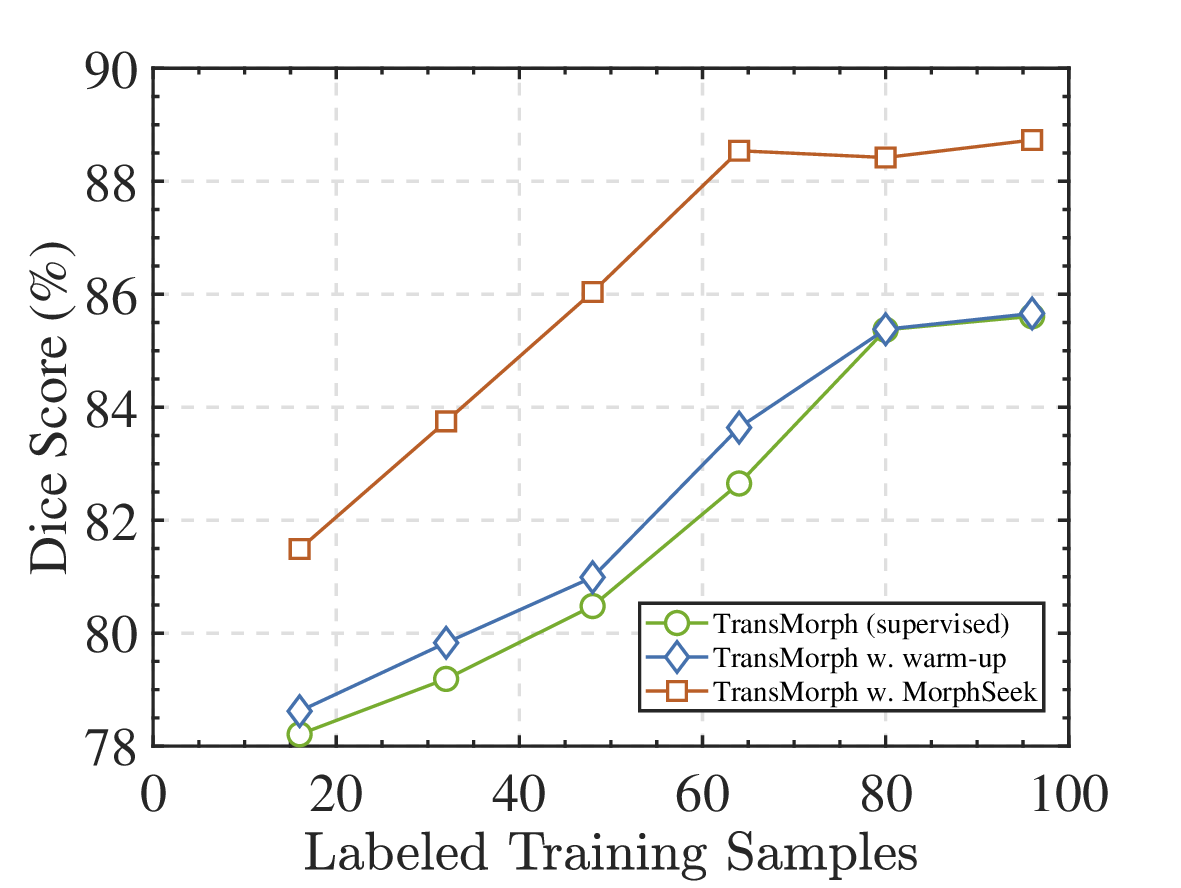}
   \caption{Impact of Warm-up and MorphSeek on GRPO Fine-tuning Performance with Limited Labeled Data (OASIS dataset)}
   \label{04}
\end{figure}

MorphSeek is particularly advantageous in weakly supervised settings with very limited labels (Fig.~\ref{04}). Using TransMorph as the backbone, MorphSeek already achieves strong gains with only about 16 labeled pairs and approaches its full-label performance with roughly 60 pairs. Notably, MorphSeek achieves 98.5\% of its full-label performance using only 60\% of the training data, while the baseline TransMorph requires 80\% of labels to reach a comparable level.

These observations support our interpretation that the multi-trajectory, multi-step GRPO scheme effectively reuses each labeled pair multiple times along the refinement steps, substantially improving the label efficiency of weak supervision and raising the performance ceiling in complex registration tasks.

\begin{table*}[!ht]
\centering
\caption{Ablation analysis of MorphSeek components on OASIS dataset.}
\label{tab:ablation_analysis}
\scalebox{0.75}{%
\begin{tabular}{clccc|cc|cc|cc}
\toprule
\multirow{2}{*}{\textbf{\#}} & \multirow{2}{*}{\textbf{Configuration}} & \multirow{2}{*}{\makecell{\textbf{Sample}\\\textbf{encoder $\mathbf{f_L}$? }}} & \multirow{2}{*}{\makecell{\textbf{Weak} \\ \textbf{Supervision?}}} & \multirow{2}{*}{\makecell{\textbf{Step}/\\\textbf{Traj}}} & \multicolumn{2}{c}{\textbf{VoxelMorph-L}} & \multicolumn{2}{c}{\textbf{TransMorph}} & \multicolumn{2}{c}{\textbf{NICE-Trans}} \\
\cmidrule(lr){6-7} \cmidrule(lr){8-9} \cmidrule(lr){10-11}
& & & & & Mean Dice (\%) $\uparrow$ & NJD (\%) $\downarrow$ & Mean Dice (\%) $\uparrow$ & NJD (\%) $\downarrow$ & Mean Dice (\%) $\uparrow$ & NJD (\%) $\downarrow$ \\
\midrule
1 & Baseline & \XSolidBrush & \XSolidBrush & 1/-- & 75.31$\pm$3.76 & 0.09$\pm$0.03 & 76.84$\pm$3.58 & 0.12$\pm$0.04 & 80.03$\pm$2.19 & 0.04$\pm$0.01 \\
2 & + Gaussian head & \checkmark & \XSolidBrush & 1/-- & 75.64$\pm$3.69 & 0.09$\pm$0.02 & 76.79$\pm$3.69 & 0.12$\pm$0.04 & 80.20$\pm$2.37 & 0.04$\pm$0.01 \\
3 & + Dice loss & \checkmark & \checkmark & 1/-- & 84.87$\pm$2.01 & 0.32$\pm$0.10 & 86.08$\pm$1.67 & 0.29$\pm$0.14 & 86.81$\pm$1.99 & 0.21$\pm$0.08 \\
4 & + Multi-step & \checkmark & \checkmark & 3/-- & 85.50$\pm$2.33 & 0.37$\pm$0.13 & 86.37$\pm$1.39 & 0.35$\pm$0.15 & 87.06$\pm$1.82 & 0.23$\pm$0.09 \\
5 & + GRPO (full) & \checkmark & \checkmark & 3/6 & \cellcolor{blue!10}\textbf{87.16$\pm$1.97} & \cellcolor{blue!10}\textbf{0.10$\pm$0.02} & \cellcolor{blue!10}\textbf{88.89$\pm$1.82} & \cellcolor{blue!10}\textbf{0.06$\pm$0.02} & \cellcolor{blue!10}\textbf{89.02$\pm$1.45} & \cellcolor{blue!10}\textbf{0.02$\pm$0.01} \\
\bottomrule
\end{tabular}%
}
\end{table*}

\subsection{Computational Overhead and the Role of Warm-up}

MorphSeek introduces less than 3\% additional parameters across all three backbones, and single-step inference latency remains close to the original models. Multi-step inference scales approximately linearly with the number of refinement steps, providing a simple deployment-time trade-off between accuracy and runtime; detailed measurements are reported in the supplement.

The training curves on OASIS (Figure~\ref{05}) highlight the critical role of unsupervised warm-up. The blue curve denotes unsupervised warm-up, whereas the green curve denotes a supervised Dice baseline using 100 labeled pairs; the dotted line marks the switch from warm-up to GRPO, so the two curves are not expected to coincide before policy optimization starts. Without warm-up, GRPO training is prone to oscillating policy gradients, higher sensitivity to hyperparameters, and a greater risk of non-physical deformations. In 20 independent runs (TransMorph backbone), warm-up increases stable-training success rate from 33\% to 79\% and reduces the average convergence epoch from approximately 120 to 75, while also producing smoother validation curves. Supplementary posterior-collapse analysis further shows that unstable runs can become nearly deterministic: sampling the same input ten times yields near-zero Dice variance after collapse, whereas normal warm-started checkpoints retain non-trivial output variance. This confirms that warm-up not only accelerates convergence but also preserves the stochastic exploration capacity required by GRPO.

\begin{figure}[t]
  \centering
  %\fbox{\rule{0pt}{2in} \rule{0.8\linewidth}{0pt}}
   \includegraphics[width=0.75\linewidth]{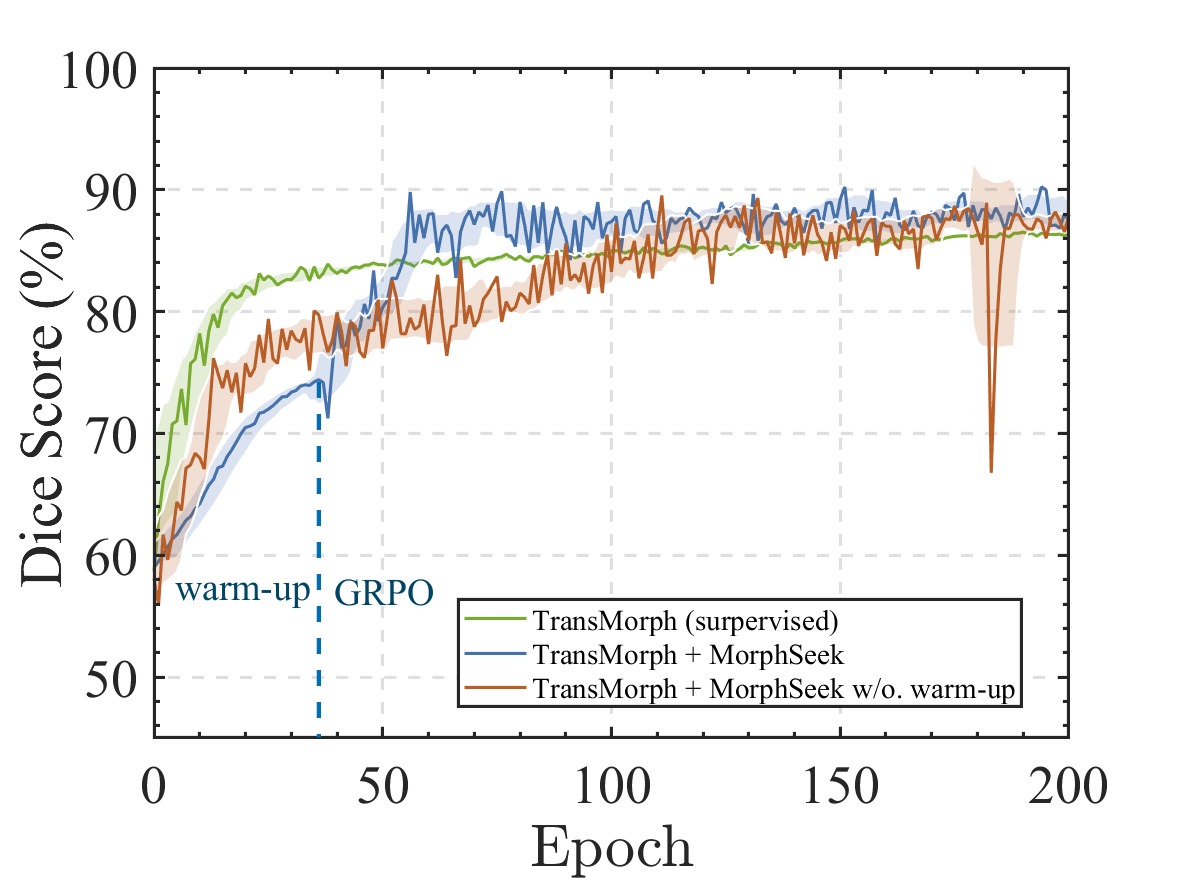}
   \caption{Validation Dice on OASIS with the TransMorph backbone. The blue curve denotes unsupervised warm-up, the green curve denotes a supervised Dice baseline using 100 labeled pairs, and the dotted line marks the switch from warm-up to GRPO.}
   \label{05}
\end{figure}

We therefore position warm-up as a prior-shaping and cost-reduction stage: it does not necessarily raise the ultimate performance ceiling, but substantially reduces the time, computational resources, and instability risks required to reach a given accuracy level, by pre-aligning the latent space before policy optimization. Supplementary failure cases further show that removing the similarity term from $\mathcal{L}_{\text{warm}}$ can yield seemingly competitive proxy metrics yet visibly smeared anatomy, reinforcing its role as an anatomy-preserving prior.

\subsection{Independent and Synergistic Contributions of MorphSeek Components}

The component-wise ablation on OASIS (Table~\ref{tab:ablation_analysis}) clarifies the contribution of each part of MorphSeek. Adding only the Gaussian head barely changes performance, validating the lightweight nature of the RL-friendly refactoring. Introducing weakly supervised Dice loss significantly boosts Dice but has limited impact on NJD, indicating that, without high-dimensional policy optimization, the available supervision signal is not fully exploited.

The full MorphSeek configuration—Gaussian head, multi-trajectory multi-step GRPO, and LDVN—achieves simultaneous improvements in both Dice and NJD across all three backbones. This demonstrates that GRPO is the key mechanism that tightly couples weak supervision with multi-step registration. In particular, the combination of latent-space policy modeling, LDVN, and multi-step GRPO yields a stable and efficient optimization scheme that lifts the performance ceiling of deformable registration.

Across tasks, modalities, and architectures, MorphSeek delivers systematic quantitative gains, with Dice improvements on the order of 2–4\% and NJD reductions of roughly 30–60\%, while also exhibiting clear advantages under low-label and resource-constrained settings. These results establish latent-space policy optimization as a practical and effective paradigm for 3D dense deformable registration.

\section{Conclusion}
\label{sec:con}

We have presented MorphSeek, which reframes deformable image registration as latent-space policy optimization and stabilizes high-dimensional GRPO through Latent-Dimension Variance Normalization (LDVN). By shifting exploration from voxel-level deformation fields to a structured latent space, MorphSeek overcomes the dimensionality and computational bottlenecks that have limited RL-based registration to low-dimensional rigid transforms. Combined with unsupervised warm-up and multi-trajectory, multi-step GRPO refinement, it consistently improves Dice while reducing NJD across three 3D benchmarks and multiple backbones, with only marginal parameter and runtime overhead, making RL-based deformable registration practical under realistic memory and label budgets. Future work includes adaptive scheduling of refinement depth, incorporating stronger physical priors on deformations, and extending latent-space policy optimization to other dense correspondence problems.

{
    \small
    \bibliographystyle{ieeenat_fullname}
    \bibliography{main}
}

\clearpage
\setcounter{page}{1}
\maketitlesupplementary

\section{Correction Note}

In previously released versions of this manuscript, a data transcription and table-formatting error occurred during the manual preparation of Table~1. As a result, some baseline entries were displayed with inconsistent rounding and unsupported least significant digits. We have re-checked the corresponding raw inference outputs and updated the affected entries to use strictly rounded one-decimal-place values.

The affected changes are limited to Table~1 in the main paper, the RIIR row in Supplementary Table~7, and the Original/SPAC row in Supplementary Table~9. These corrections do not change the relative performance trends, method ranking, or any scientific conclusions of the paper. The higher-precision values reported in the ablation and supplementary analyses were independently verified and are unaffected by this correction. We apologize for any confusion caused.

\section{Analysis of Latent-Dimension Variance Normalization (LDVN)}
\label{sec:ldvn}

We reuse the notation in Sec.~\ref{subsec:grpo}. LDVN modifies the log-likelihood term by introducing a latent-dimension-aware scaling.
We define the LDVN-transformed log-likelihood as
\begin{equation}
  \hat{\ell}^{(j)}
  \;\triangleq\;
  \frac{1}{s}\,\log \tilde{\pi}^{(j)}
  =
  \frac{1}{s} \big( \log \pi^{(j)} - \log \bar{\pi} \big),
  \label{eq:ldvn_def}
\end{equation}
where $s>0$ is a scaling factor that depends on the latent dimensionality $N$ (specified in Sec.~\ref{sec:ldvn_var_scaling}).
The LDVN-based policy loss is then
\begin{equation}
  \mathcal{L}_\text{policy}^{\text{LDVN}}(\theta_E)
  = -\frac{1}{J} \sum_{j=1}^J A^{(j)}\,\hat{\ell}^{(j)}.
  \label{eq:policy_loss_ldvn}
\end{equation}

\paragraph{Affine invariance under zero-mean advantages.}
We first show that LDVN does not alter the underlying optimization objective: it only rescales the gradient magnitude while preserving its direction and fixed points.

Consider the more general affine form
\begin{equation}
  \hat{\ell}^{(j)}
  \;=\; 
  \alpha \,\log \pi^{(j)} 
  + \beta \,\log \bar{\pi} 
  + b,
  \quad
  \alpha > 0,\; \beta, b \in \mathbb{R},
  \label{eq:affine_logpi}
\end{equation}
and the corresponding policy loss
\begin{equation}
  \mathcal{L}_\text{policy}^{\text{affine}}(\theta_E)
  = -\frac{1}{J} \sum_{j=1}^J A^{(j)}\,\hat{\ell}^{(j)}.
  \label{eq:policy_loss_affine}
\end{equation}

\medskip\noindent\textbf{Proposition 1.}
\emph{
Under the zero-mean advantage condition in Eq.~\ref{A_j},
the gradient of $\mathcal{L}_\text{policy}^{\text{affine}}$ with respect to the encoder parameters $\theta_E$ is
\[
\nabla_{\theta_E} \mathcal{L}_\text{policy}^{\text{affine}}
= -\frac{\alpha}{J} \sum_{j=1}^J A^{(j)}\,\nabla_{\theta_E} \log \pi^{(j)}.
\]
In particular, any affine transform of the form \ref{eq:affine_logpi} preserves the policy-gradient direction and only rescales its magnitude by the positive constant $\alpha$.
}

\medskip\noindent\emph{Proof.}
Since $b$ does not depend on $\theta_E$, we have
\[
\nabla_{\theta_E} \hat{\ell}^{(j)}
= \alpha \,\nabla_{\theta_E} \log \pi^{(j)} 
  + \beta \,\nabla_{\theta_E} \log \bar{\pi}.
\]
Using the definition of $\log \bar{\pi}$,
\[
\nabla_{\theta_E} \log \bar{\pi}
= \nabla_{\theta_E} \frac{1}{J} \sum_{k=1}^J \log \pi^{(k)}
= \frac{1}{J} \sum_{k=1}^J \nabla_{\theta_E} \log \pi^{(k)}.
\]
Therefore,
\begin{align*}
\nabla_{\theta_E} \mathcal{L}_\text{policy}^{\text{affine}}
&= -\frac{1}{J} \sum_{j=1}^J A^{(j)} 
   \left[\alpha \,\nabla_{\theta_E} \log \pi^{(j)} 
        + \beta \,\nabla_{\theta_E} \log \bar{\pi} \right] \\
&= -\frac{\alpha}{J} \sum_{j=1}^J A^{(j)} \nabla_{\theta_E} \log \pi^{(j)}
   - \\ &\frac{\beta}{J} \left(\sum_{j=1}^J A^{(j)}\right) 
                    \left(\frac{1}{J}\sum_{k=1}^J \nabla_{\theta_E} \log \pi^{(k)}\right).
\end{align*}
By Eq.~\ref{A_j}, $\sum_{j=1}^J A^{(j)} = 0$, hence the second term vanishes exactly and we obtain
\[
\nabla_{\theta_E} \mathcal{L}_\text{policy}^{\text{affine}}
= -\frac{\alpha}{J} \sum_{j=1}^J A^{(j)} \nabla_{\theta_E} \log \pi^{(j)}.
\]
Thus the gradient direction coincides with the standard GRPO gradient, up to a global positive scalar $\alpha$, proving the claim.

Taking $\alpha = 1/s$ and $\beta = -1/s$ recovers LDVN in Eq.~\ref{eq:ldvn_def}. Thus LDVN does not change gradient direction or fixed points, and only adjusts the effective update scale.

\subsection{Dimension-dependent variance and choice of \texorpdfstring{$s$}{s}}
\label{sec:ldvn_var_scaling}

We now analyze how the variance of the log-likelihood grows with the latent dimensionality $N$ and use this to derive a principled choice for the scaling factor $s$.

For the Gaussian policy in Eq.~\ref{warmup} of the main paper, the log-likelihood of a sampled latent code
$\mathbf{z} = (z_1,\dots,z_N)$ can be written as a sum of $N$ per-dimension contributions:
\begin{equation}
\begin{split}
 & \log \pi(\mathbf{z} \mid \boldsymbol{\mu}, \boldsymbol{\sigma}) \\
  &= -\frac{1}{2} \sum_{i=1}^N 
    \left[
      \left(\frac{z_i - \mu_i}{\tau\sigma_i}\right)^2 
      + \log(2\pi \tau^2 \sigma_i^2)
    \right] \triangleq
  \sum_{i=1}^N X_i,   
\end{split}
\label{eq:logpi_sum}
\end{equation}
where $X_i$ denotes the contribution from the $i$-th latent dimension.

We assume that the per-dimension terms $\{X_i\}$ have uniformly bounded second moments and are at most weakly dependent.
Under these mild conditions, the variance of the sum in Eq.~\ref{eq:logpi_sum} satisfies
\begin{align}
  \mathrm{Var}&\Big[\log \pi(\mathbf{z} \mid \boldsymbol{\mu}, \boldsymbol{\sigma})\Big]
  = \mathrm{Var}\Big[\sum_{i=1}^N X_i\Big] \nonumber\\
  &= \sum_{i=1}^N \mathrm{Var}(X_i)+ 2\sum_{1 \le i < k \le N} \mathrm{Cov}(X_i,X_k).
  \label{eq:var_sum_expansion}
\end{align}
If $\mathrm{Var}(X_i) \le C$ for all $i$ and the covariance terms are either zero or sufficiently sparse/decaying, the right-hand side grows at most linearly in $N$, so
\begin{equation}
  \mathrm{std}\big[\log \pi(\mathbf{z} \mid \boldsymbol{\mu}, \boldsymbol{\sigma})\big]
  = O(\sqrt{N}).
  \label{eq:var_logpi_on}
\end{equation}

Subtracting the group mean does not change the order of magnitude, so $\mathrm{std}(\log \tilde{\pi}^{(j)}) = O(\sqrt{N})$.

Moreover, for any $s>0$ and $b\in\mathbb{R}$,
\begin{equation}
\left(\frac{1}{s}\log\pi^{(j)}+b\right)-\overline{\left(\frac{1}{s}\log\pi+b\right)}=\frac{1}{s}\left(\log\pi^{(j)}-\overline{\log\pi}\right).
\end{equation}
This shows that LDVN is an affine, dimension-aware transformation of the group-relative log-likelihood: it preserves within-group ordering and the policy-gradient direction while only rescaling update magnitude.

Given Eq.~\ref{eq:var_logpi_on}, we now choose $s$ such that the variance of the LDVN-transformed log-likelihood $\hat{\ell}^{(j)}$ in Eq.~\ref{eq:ldvn_def} remains stable as $N$ grows.
Using the fact that scaling a random variable by $1/s$ divides its variance by $s^2$, we obtain
\begin{equation}
  \mathrm{Var}\big[\hat{\ell}^{(j)}\big]
  =  \frac{1}{s^2}\,\mathrm{Var}\big[\log \tilde{\pi}^{(j)}\big]
  = \frac{1}{s^2}\,O(N).
  \label{eq:var_ldvn}
\end{equation}
To make this variance $O(1)$, independent of the latent dimensionality, we require
\[
\frac{N}{s^2} = O(1)
\quad\Longrightarrow\quad
s^2 \propto N 
\quad\Longrightarrow\quad
s \propto \sqrt{N}.
\]

The above derivation is mathematically analogous to the scaled dot-product attention used in Transformers\cite{0402attentionisallyouneed}, where the dot product between query and key vectors is divided by $\sqrt{d_k}$ to prevent its variance from growing with the feature dimension $d_k$.
Here, LDVN plays the same role for log-likelihoods in high-dimensional latent spaces: by normalizing $\log \tilde{\pi}^{(j)}$ with $1/\sqrt{N}$, we keep its variance roughly constant across different latent dimensionalities, stabilizing GRPO updates without additional parameters.

In summary, LDVN applies an affine, dimension-aware transformation to the group-relative log-likelihood that (i) preserves policy-gradient direction while rescaling magnitude and (ii) cancels the $O(\sqrt{N})$ growth of its standard deviation.

This turns latent-space policy optimization into a numerically stable procedure even under high-dimensional latent codes, which is crucial for making RL-based registration practically viable beyond low-dimensional rigid transformations.

\subsection{Ablation Studies for LDVN}
\label{sec:ldvn_ablation}

\begin{figure}[t]
  \centering
  \includegraphics[width=0.75\linewidth]{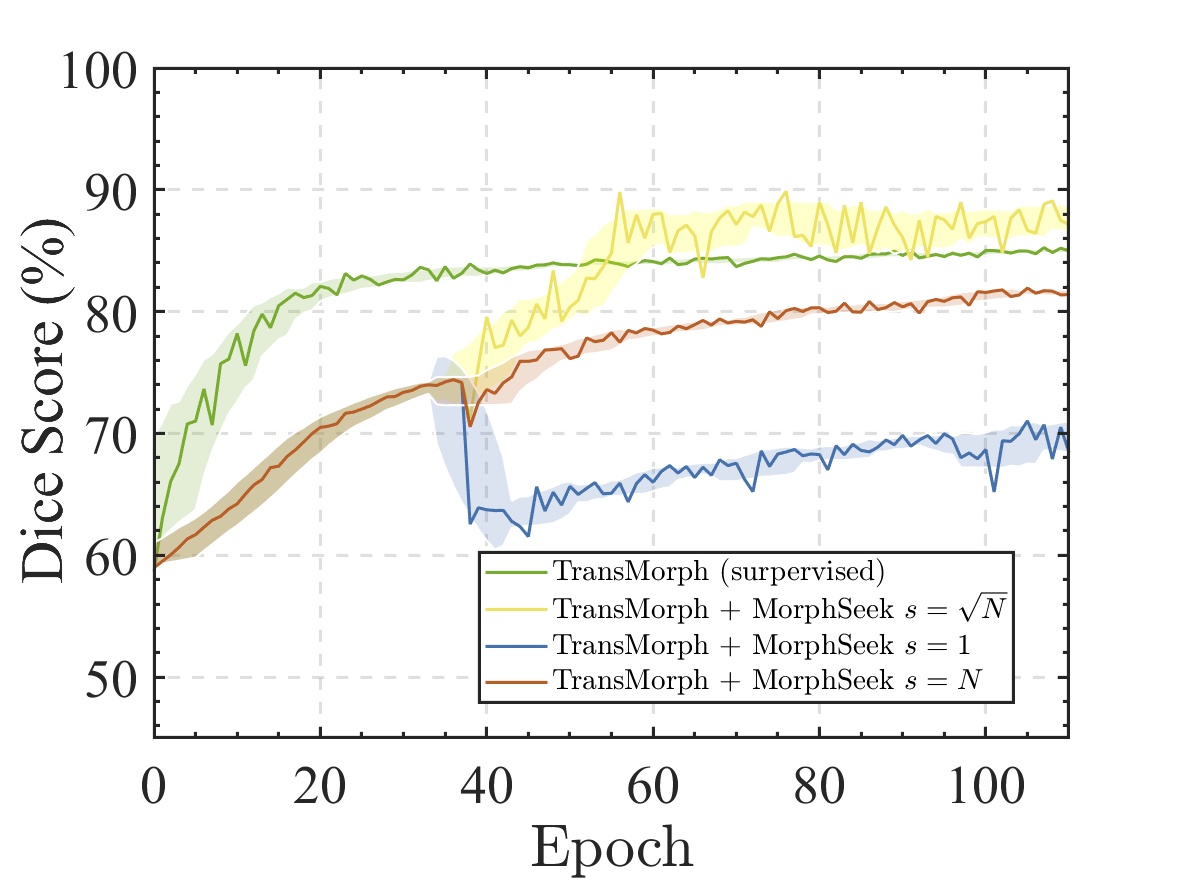}
  \caption{
    Validation Dice on OASIS for TransMorph under different LDVN scaling
    factors $s$.
  }
  \label{fig:ldvn_curve}
\end{figure}

We revisit the Gaussian policy log-likelihood with the LDVN scaling factor
$s$ (Eq.~\ref{logpi}) and ablate different choices of $s$. On the TransMorph+OASIS task, we keep all settings identical to the main
paper and vary only
the LDVN scaling factor,
\[
s \in \{ 1, \sqrt{N}, N \}.
\]
We also include a purely supervised TransMorph baseline trained with the
Dice loss.
Figure~\ref{fig:ldvn_curve} reports validation Dice scores over training
epochs.

As a result, when $s = N$, the GRPO contribution to the
loss is weak; the curve almost coincides with the supervised baseline.
When $s = 1$, the variance of $\log \tilde{\pi}^{(j)}$ grows with $N$,
GRPO gradients become noisy, and the model forgets the warm-up
representation; the final Dice remains below both the baseline and the other
settings. With $s = \sqrt{N}$, the variance is stabilized at $\mathcal{O}(1)$,
GRPO updates are stable.

These observations empirically support the choice $s \propto \sqrt{N}$ for
high-dimensional latent policies.

\subsection{Empirical Check of the Weak-Dependence Assumption}
\label{sec:weak_dep_empirical}

To complement the variance derivation in Sec.~\ref{sec:ldvn_var_scaling},
we empirically examine the weak-dependence assumption on OASIS with the
TransMorph backbone ($N\approx1.6\times10^5$).
For each checkpoint, we draw $10^3$ Monte Carlo latent samples and estimate
\(\mathrm{Var}(\sum_i X_i)\), where $X_i$ is the per-dimension contribution in
Eq.~\ref{eq:logpi_sum}.

We report the ratio between empirical variance and the independence baseline
(\(0.5N\)). The observed ratio is
\(1.01\pm0.04\), indicating negligible cross-dimension correlation in practice.
This supports the \(O(N)\) variance growth assumption and the choice
\(s\propto\sqrt{N}\).

\begin{table}[t]
\centering
\caption{Monte Carlo verification of weak dependence on OASIS (TransMorph, $10^3$ samples).}
\label{tab:weak_dep_mc}
\begin{tabular}{lc}
\toprule
Metric & Value \\
\midrule
$\mathrm{Var}_{emp}(\sum_i X_i)/(0.5N)$ & $1.01\pm0.04$ \\
\bottomrule
\end{tabular}
\end{table}

\subsection{Critical Hyperparameter Sensitivity}
\label{sec:hyper_sensitivity}

We analyze key stability-related hyperparameters on OASIS with TransMorph
(50 GRPO epochs). Table~\ref{tab:hyper_supp} summarizes representative
failure modes when deviating from the default setting.

\begin{table}[t]
\centering
\caption{Hyperparameter sensitivity on OASIS (TransMorph).}
\label{tab:hyper_supp}
\resizebox{\linewidth}{!}{
\begin{tabular}{lcc}
\toprule
Setting & Dice$\uparrow$/NJD$\downarrow$ & Observation \\
\midrule
Default & 88.44/0.06 & Stable \\
$\tau=1$ & 83.13/0.04 & Under-exploration \\
$\tau=15$ & 55.67/0.01 & Exploration collapse \\
No $\sigma$ clip & 33.05/3.84 & Instability \\
$\sigma_{\max}=0$ & 83.87/0.10 & Under-exploration \\
$\lambda_{\mathrm{KL}}=0.1$ & 86.20/0.22 & Posterior collapse tendency \\
$\lambda_{\mathrm{KL}}=0$ & 49.63/0.00 & Collapse \\
$\omega_{\mathrm{Dice}}=0$ & 82.99/0.10 & Weaker alignment \\
$\omega_{\mathrm{NJD}}=0$ & 89.52/0.59 & Poor regularity \\
\bottomrule
\end{tabular}}
\end{table}

In addition, over 20 independent runs, warm-up improves the stable-training
success rate from 33\% to 79\% and reduces the convergence epoch from
approximately 120 to 75.

\subsection{Posterior Collapse Analysis}
\label{sec:collapse_supp}

We additionally probe posterior collapse by sampling the latent code ten times for
the same input at representative checkpoints of VoxelMorph-L on OASIS. Table~\ref{tab:collapse_supp}
reports the mean and standard deviation of Dice across the ten samples. Near-zero
standard deviation indicates that the encoder has become almost deterministic,
removing the exploration signal required by GRPO. This analysis empirically
motivates the deterministic warm-up in Eq.~\ref{z}: initializing the model on
the mean code before stochastic sampling reduces collapse risk and helps preserve
non-trivial output variance. Without warm-up, or after unstable GRPO under poor
hyperparameters, the model can drift toward this regime; the normal warm-started
checkpoint instead retains non-trivial output variance.

\begin{table}[t]
\centering
\caption{Posterior collapse analysis on OASIS. Dice is reported as mean$\pm$std (\%) over ten latent samples for the same input pair.}
\label{tab:collapse_supp}
\resizebox{\linewidth}{!}{%
\begin{tabular}{cccc}
\toprule
Ep. 0 w/o warm-up & Ep. 0 w/ warm-up & Ep. 100 (collapsed) & Ep. 100 (normal)\\
\midrule
73.26$\pm$0.05 & 69.00$\pm$3.75 & 88.89$\pm$0.00 & 90.42$\pm$0.67\\
\bottomrule
\end{tabular}}
\end{table}

\subsection{Why Keep \texorpdfstring{$\mathcal{L}_{\text{warm}}$}{L_warm} During GRPO?}
\label{sec:lwarm_supp}

During GRPO, Dice and NJD act only as proxy rewards. If the optimization is left
unconstrained, the policy may exploit these proxies by producing anatomically
implausible deformations that still look numerically acceptable. Retaining
$\mathcal{L}_{\text{warm}}$ keeps updates close to the anatomy-preserving manifold
learned during warm-up.

\begin{figure}[t]
  \centering
  \includegraphics[width=0.72\linewidth]{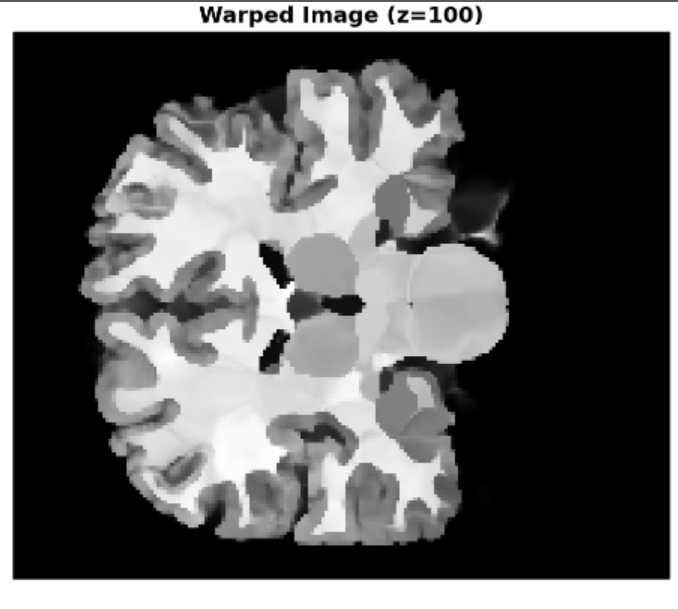}
  \caption{Failure case when removing the similarity term from $\mathcal{L}_{\text{warm}}$ during GRPO. Although proxy metrics can remain deceptively favorable, the warped anatomy becomes smeared and physically implausible, indicating reward hacking.}
  \label{fig:reward_hacking_supp}
\end{figure}

Figure~\ref{fig:reward_hacking_supp} illustrates this failure mode: without the
similarity term in $\mathcal{L}_{\text{warm}}$, GRPO can drive the deformation toward
medically meaningless structures that artificially improve overlap-oriented rewards.
This observation motivates keeping the warm-up objective during policy optimization,
rather than treating it as a pure initialization stage.

\subsection{Additional Comparison with Multi-stage Baselines}
\label{sec:lapirn_supp}

Table~\ref{tab:lapirn_supp} compares MorphSeek with representative step-wise or
cascaded alternatives on OASIS. RIIR is already included in the main paper; we
add LapIRN-stage3 here because it is another canonical multi-stage registration
baseline. Under the same 100-pair labeled setting, MorphSeek achieves the best
overall trade-off between accuracy and deformation regularity.

\begin{table}[t]
    \centering
    \caption{Additional comparison with multi-stage baselines on OASIS. LapIRN is trained with the same 100 labeled pairs used in our weakly supervised setting.}
    \label{tab:lapirn_supp}
    \resizebox{\linewidth}{!}{%
        \begin{tabular}{lcc}
        \toprule
        Method & Dice$\uparrow$(\%) & NJD$\downarrow$(\%) \\
        \midrule
        RIIR (12 steps) & 87.7$\pm$2.5 & 0.1$\pm$0.0 \\
        LapIRN-diff (stage3) & 79.70$\pm$2.96 & 0.09$\pm$0.03 \\
        LapIRN-disp (stage3) & 84.52$\pm$1.64 & 3.13$\pm$0.41 \\
        TransMorph + MorphSeek (3/6) & \textbf{88.89$\pm$1.82} & \textbf{0.06$\pm$0.02} \\
        \bottomrule
        \end{tabular}%
    }
\end{table}

\section{Classical Optimization-Based Baselines}
\label{sec:classical_baselines}

\begin{table*}[ht]
    \centering
    \caption{Performance of classical optimization-based baselines on the three benchmarks. We report mean Dice [\%] (higher is better), NJD [\%] (lower is better), and CPU time per test pair in seconds (lower is better).}
    \label{tab:classical_baselines}
    \setlength{\tabcolsep}{4pt}
    \scalebox{0.8}{%
    \begin{tabular}{lccccccccc}
        \toprule
        \multirow{2}{*}{Method} 
        & \multicolumn{3}{c}{Dice [\%] $\uparrow$} 
        & \multicolumn{3}{c}{NJD [\%] $\downarrow$} 
        & \multicolumn{3}{c}{CPU time [s] $\downarrow$} \\
        \cmidrule(lr){2-4}\cmidrule(lr){5-7}\cmidrule(lr){8-10}
        & OASIS & LiTS & AbMRCT 
        & OASIS & LiTS & AbMRCT 
        & OASIS & LiTS & AbMRCT \\
        \midrule
        SyN 
        & $75.53\pm3.29$ & $79.13\pm11.26$ & $44.28\pm28.79$ 
        & $0.00\pm0.00$  & $0.00\pm0.00$   & $0.01\pm0.00$ 
        & $48.66\pm0.00$ & $47.47\pm0.00$  & $47.01\pm0.00$ \\
        deedsBCV 
        & $76.38\pm2.89$ & $77.14\pm17.77$ & $58.99\pm20.31$ 
        & $0.23\pm0.15$  & $0.19\pm0.12$   & $0.25\pm0.19$ 
        & $33.18\pm0.00$ & $31.52\pm0.02$  & $30.08\pm0.08$ \\
        \bottomrule
    \end{tabular}}
\end{table*}

For completeness and to contextualize our learning-based results against strong optimization-based methods, we additionally evaluate two classical non–deep-learning registration algorithms on the same test splits and registration directions as in the main paper. Specifically, we consider SyN from ANTs~\cite{0201}, a classical standard in the field, and deedsBCV~\cite{beed}, a more recent method based on discrete optimization.

For SyN, we follow common practice and use normalized cross-correlation (\texttt{syn\_metric=CC}) on the mono-modality OASIS and LiTS datasets, and Mattes mutual information (\texttt{syn\_metric=mattes})~\cite{mattes2001mi} on the cross-modality Abdomen MR$\leftarrow$CT (AbMRCT) task. Across all three benchmarks, the multi-resolution schedule is set to \texttt{reg\_iterations=(60, 40, 20)}, with all remaining parameters kept at their default values. 

For deedsBCV, we use self-similarity context (SSC) as the objective function on all datasets. On OASIS, the grid-spacing, search-radius, and quantization-step pyramids are set to $6\times5\times4\times3\times2$, $6\times5\times4\times3\times2$, and $5\times4\times3\times2\times1$, respectively; on LiTS and AbMRCT, the corresponding settings are $8\times7\times6\times5\times4$, $8\times7\times6\times5\times4$, and $5\times4\times3\times2\times1$.

Table~\ref{tab:classical_baselines} summarizes the resulting mean Dice [\%], NJD [\%], and per-pair CPU time [s] over the test sets. Classical methods remain competitive on OASIS and LiTS but degrade notably on the more challenging AbMRCT task, and they require tens of seconds per case, highlighting the computational overhead of purely optimization-based registration compared with learning-based approaches discussed in the main paper.

\section{Discussions: Why MorphSeek Enables Reliable NJD While SPAC Does Not? }

A central design choice in MorphSeek is to make the
multi-step refinement fully traceable on a \emph{fixed} reference grid.
At each step $t$, MorphSeek maintains an explicit cumulative
deformation field $\Phi_t$ and updates it by composing the
incremental displacement $\varphi_t$ predicted at that step:
\begin{equation}
\Phi_t = \varphi_t \circ \Phi_{t-1}.
\label{eq:phi_recursive}
\end{equation}
Both the loss terms and the warped image $I_m \circ \Phi_t$
are computed using this composed field. Consequently, the
final deformation $\Phi_T$ is \emph{exactly} the field that produces
the reported $I_{\text{warped}} = I_m \circ \Phi_T$, and the NJD metric
can be directly evaluated on the same deformation that is
responsible for the quantitative results in Table~\ref{tab:main_results}.
This explicit accumulation makes NJD a well-defined and
reproducible measure of deformation regularity for MorphSeek
and all refactored baselines.

When we attempted to apply the same NJD protocol to the
RL-based SPAC framework, we encountered a structural
mismatch between its inference scheme and the requirements
for reliable Jacobian analysis. Although SPAC and MorphSeek
both adopt multi-step refinement, SPAC does \emph{not} maintain
a cumulative deformation field on the original coordinate
system. Instead, each predicted single-step displacement is
applied directly to the current moving image $I_m^t$, and only
the intermediate images $I_m^t$ and per-step fields $\phi_t$
are stored. Conceptually, the final warped image can be
written as
\begin{equation}
I_{\text{warped}}
=
I_m \circ \phi_T \circ \phi_{T-1} \circ \dots \circ \phi_1,
\label{eq:spac_composition_ideal}
\end{equation}
where $\phi_t$ denotes the displacement predicted at step $t$
in the current image coordinates. However, during inference
SPAC does not construct or output the exact total deformation
$\Phi_T$ that maps the original $I_m$ to $I_{\text{warped}}$ on a fixed grid.

To make NJD computation possible for SPAC, one must
therefore reconstruct a ``total'' deformation post hoc by
composing the saved $\{\phi_t\}_{t=1}^{T}$ via displacement
composition, e.g., using standard ITK-style operators.
This inevitably introduces several sources of numerical
inconsistency that MorphSeek deliberately avoids by
operating on a single reference grid:
\begin{itemize}
  \item \textbf{Interpolation error.}
        Each resampling of a deformation field smooths the
        displacement vectors and introduces small geometric
        deviations; repeating this over many steps amplifies
        the discrepancy between the reconstructed field and
        the effective transformation applied during inference.
  \item \textbf{Discretization error.}
        When the deformation varies rapidly within a voxel,
        a single sampled displacement cannot faithfully
        represent the local transformation, leading to biased
        Jacobian estimates once fields are repeatedly regridded.
  \item \textbf{Non-associativity at the discrete level.}
        In continuous space, composition is associative,
        $((\phi_3 \circ \phi_2) \circ \phi_1) = \phi_3 \circ (\phi_2 \circ \phi_1)$.
        Under ``interpolation + grid truncation'', different
        composition orders yield slightly different discrete
        fields, and these differences accumulate across many
        refinement steps.
\end{itemize}

In practice, SPAC often uses on the order of $T \approx 20$
refinement steps. After composing $\{\phi_t\}_{t=1}^{T}$ into an
approximate total field $\hat{\Phi}_T$ using several reasonable
composition schemes, we observe that warping $I_m$ with
$\hat{\Phi}_T$ yields segmentations whose Dice scores are more
than 10\% worse than those obtained from the original SPAC
inference $I_{\text{warped}}$. In other words, the reconstructed
$\hat{\Phi}_T$ no longer reproduces the reported SPAC behaviour,
so any NJD computed on $\hat{\Phi}_T$ would characterize a
different, numerically degraded deformation.

\begin{table}[t]
\centering
\caption{Effect of different post-hoc composition schemes on SPAC. Dice and NJD are computed
using the reconstructed total deformation $\hat{\Phi}_T$ rather than the original SPAC output (OASIS task).}
\label{tab:spac_composition}
\begin{tabular}{lcc}
\toprule
Composition scheme & Dice (\%) & NJD (\%) \\
\midrule
Original                        &   $78.9\pm 5.3$    &   N/A   \\
Vector summation                           &   $61.27\pm 7.86$    &  $1.42 \pm 0.41$    \\
\begin{tabular}[c]{@{}c@{}}Displacement composition \\ + trilinear interp\end{tabular} &   $66.35 \pm 4.39$    &  $0. 25\pm 0.17$   \\
\begin{tabular}[c]{@{}c@{}}Displacement composition \\ + B-spline interp\end{tabular}  &    $68.09 \pm 6.44$   &   $0.30\pm0.20$   \\
\bottomrule
\end{tabular}
\end{table}

Table~\ref{tab:spac_composition} summarizes this effect on
the OASIS task: all post-hoc composition schemes lead to
substantial Dice drops and inconsistent NJD values when
evaluated on $\hat{\Phi}_T$. These observations indicate that
NJD cannot be reliably reported for SPAC without
redefining its inference pipeline and output representation.
For this reason, we refrain from listing NJD for SPAC in
our experiments. In contrast, MorphSeek and the refactored
U-Net baselines are expressly designed to maintain an
explicit cumulative $\Phi_T$ on a fixed grid, ensuring that
the reported NJD always reflects the \emph{actual} deformation
that produced the corresponding registration results.

\section{Implementation Details and Reproducibility}
\label{sec:impl}

\subsection{Loss Definitions and Similarity Metrics}
\label{sec:loss}

The unsupervised warm-up loss in Eq.~\ref{warmup} of the main paper is
\begin{equation}
\mathcal{L}_{\text{warm}}(\boldsymbol{\theta})
= \mathcal{L}_{\text{sim}}(I_f, I_m \circ \Phi)
+ \lambda_{\text{reg}} \mathcal{L}_{\text{reg}}(\Phi)
+ \beta_{\text{KL}} \mathcal{L}_{\text{KL}},
\end{equation}
where $I_f, I_m : \Omega \rightarrow \mathbb{R}$ are fixed and moving images on voxel grid $\Omega$, and $\Phi : \Omega \rightarrow \mathbb{R}^3$ is the predicted deformation. We denote the warped image by $\tilde{I}_m = I_m \circ \Phi$ and use a cubic window $\mathcal{N}(\mathbf{p})$ of side length $w=9$ centered at voxel $\mathbf{p}$ for windowed quantities.

\paragraph{Image similarity.}
For OASIS and LiTS we use a local MSE similarity:
\begin{equation}
\mathcal{L}^{\text{MSE}}_{\text{sim}}(I_f, I_m \circ \Phi)
= \frac{1}{|\Omega|}
\sum_{\mathbf{p} \in \Omega}
\frac{1}{|\mathcal{N}(\mathbf{p})|}
\sum_{\mathbf{q} \in \mathcal{N}(\mathbf{p})}
\big(I_f(\mathbf{q}) - \tilde{I}_m(\mathbf{q})\big)^2.
\end{equation}
For Abdomen MR$\leftarrow$CT we replace MSE with a standard implementation of the MIND descriptor~\cite{0808mindmetric}, which we use directly as $\mathcal{L}_{\text{sim}}$.

For completeness, we also consider a windowed NCC variant. Let
\begin{align}
\mu_f(\mathbf{p}) &= \frac{1}{|\mathcal{N}(\mathbf{p})|}
\sum_{\mathbf{q} \in \mathcal{N}(\mathbf{p})} I_f(\mathbf{q}), \\
\mu_m(\mathbf{p}) &= \frac{1}{|\mathcal{N}(\mathbf{p})|}
\sum_{\mathbf{q} \in \mathcal{N}(\mathbf{p})} \tilde{I}_m(\mathbf{q}),
\end{align}
and define zero-mean patches
\(
\hat{I}_f(\mathbf{q}; \mathbf{p}) = I_f(\mathbf{q}) - \mu_f(\mathbf{p}),
\hat{I}_m(\mathbf{q}; \mathbf{p}) = \tilde{I}_m(\mathbf{q}) - \mu_m(\mathbf{p})
\).
The local NCC at $\mathbf{p}$ is
\begin{equation}
\mathrm{NCC}(\mathbf{p}) =
\frac{\sum_{\mathbf{q} \in \mathcal{N}(\mathbf{p})}
\hat{I}_f(\mathbf{q}; \mathbf{p}) \hat{I}_m(\mathbf{q}; \mathbf{p})}
{\sqrt{\sum_{\mathbf{q}} \hat{I}_f(\mathbf{q}; \mathbf{p})^2}
 \sqrt{\sum_{\mathbf{q}} \hat{I}_m(\mathbf{q}; \mathbf{p})^2} + \varepsilon},
\end{equation}
and the corresponding loss is the negative average correlation:
\begin{equation}
\mathcal{L}^{\text{NCC}}_{\text{sim}}(I_f, I_m \circ \Phi)
= -\frac{1}{|\Omega|} \sum_{\mathbf{p} \in \Omega} \mathrm{NCC}(\mathbf{p}).
\end{equation}

\paragraph{Deformation regularization.}
Let $\mathbf{u}(\mathbf{x}) = \Phi(\mathbf{x}) - \mathbf{x}$ be the displacement.
We use an $\ell_2$ diffusion penalty on first-order finite differences:
\begin{equation}
\mathcal{L}_{\text{reg}}(\Phi)
= \frac{1}{|\Omega|}
\sum_{\mathbf{p} \in \Omega}
\sum_{d \in \{x,y,z\}}
\big\| \nabla_d \mathbf{u}(\mathbf{p}) \big\|_2^2,
\end{equation}
where $\nabla_d$ denotes the discrete difference along spatial direction $d$.

\paragraph{KL penalty on Gaussian heads.}
The encoder defines a factorized Gaussian
$q_{\boldsymbol{\theta}_E}(\mathbf{z} \mid f_L)
= \mathcal{N}(\boldsymbol{\mu}, \mathrm{diag}(\boldsymbol{\sigma}^2))$
over the latent tensor $\mathbf{z} = f_L$ with $N$ total dimensions.
The KL term is the standard divergence to the unit Gaussian prior:
\begin{equation}
\mathcal{L}_{\text{KL}}
= \frac{1}{2N} \sum_{i=1}^N
\big(\mu_i^2 + \sigma_i^2 - \log \sigma_i^2 - 1\big).
\end{equation}

\subsection{Reward Shaping and Jacobian Regularity}
\label{sec:reward}

During GRPO fine-tuning we use a reward that combines hard Dice gains with a Jacobian-based regularity term, and an auxiliary soft Dice loss.

\paragraph{Hard Dice for reward shaping.}
Let $S_f, S_m : \Omega \rightarrow \{0, 1, \dots, K\}$ denote fixed and moving segmentations. For a deformation $\Phi$, we warp the moving labels by nearest-neighbor interpolation,
\begin{equation}
\tilde{S}_m(\mathbf{x}) = S_m(\Phi(\mathbf{x})), \quad \mathbf{x} \in \Omega,
\end{equation}
and derive one-hot maps $S_f^c, \tilde{S}_m^c : \Omega \rightarrow \{0,1\}$ for each class $c$.
The per-class hard Dice coefficient is
\begin{equation}
\mathrm{Dice}^{\text{hard}}_c(S_f, \tilde{S}_m)
= \frac{2 \sum_{\mathbf{x} \in \Omega} S_f^c(\mathbf{x}) \tilde{S}_m^c(\mathbf{x})}
{\sum_{\mathbf{x}} S_f^c(\mathbf{x}) + \sum_{\mathbf{x}} \tilde{S}_m^c(\mathbf{x}) + \varepsilon},
\end{equation}
and the macro-averaged multi-class Dice is
\begin{equation}
\mathrm{Dice}^{\text{hard}}(S_f, \tilde{S}_m)
= \frac{1}{K} \sum_{c=1}^K
\mathrm{Dice}^{\text{hard}}_c(S_f, \tilde{S}_m).
\end{equation}
At GRPO step $t$, each trajectory $j$ produces a deformation
$\Phi_t^{(j)}$ and Dice
$D_t^{(j)} = \mathrm{Dice}^{\text{hard}}(S_f, S_m \circ \Phi_t^{(j)})$.
With baseline deformation $\Phi_{t-1}$ and $D_{t-1}$ (identity for $t=1$), the Dice gain is
\begin{equation}
\Delta D^{(j)} = D_t^{(j)} - D_{t-1}.
\end{equation}

\paragraph{Jacobian regularity (NJD).}
For $\Phi(\mathbf{x}) = \mathbf{x} + \mathbf{u}(\mathbf{x})$ we approximate
\begin{equation}
\mathbf{J}_\Phi(\mathbf{x})
= \frac{\partial \Phi(\mathbf{x})}{\partial \mathbf{x}}
\approx \mathbf{I}_3 + \nabla \mathbf{u}(\mathbf{x}),
\end{equation}
and define the set of folding voxels
\begin{equation}
\Omega_{-}(\Phi)
= \big\{ \mathbf{x} \in \Omega : \det \mathbf{J}_\Phi(\mathbf{x}) < 0 \big\}.
\end{equation}
The negative-Jacobian determinant percentage is
\begin{equation}
\mathrm{NJD}(\Phi)
= \frac{|\Omega_{-}(\Phi)|}{|\Omega|},
\end{equation}
i.e., the fraction of voxels with local foldings.

\paragraph{Step-wise reward and soft Dice loss.}
The step-wise reward for trajectory $j$ is
\begin{equation}
R^{(j)} = w_{\text{Dice}} \, \Delta D^{(j)}
+ w_{\text{NJD}} \, \mathrm{NJD}(\Phi^{(j)}),
\end{equation}
with $w_{\text{Dice}} > 0$ and $w_{\text{NJD}} < 0$.
These rewards are group-normalized to compute advantages, which are combined with LDVN-normalized log-probabilities in the policy loss.
Compared with optimizing Dice alone, which induces a greedy and deterministic update from the current prediction, GRPO uses relative ranking over sampled trajectories and therefore provides a smoother exploration-based training signal. In practice, evaluating multiple hypotheses per pair helps smooth the highly non-convex registration landscape and makes it easier to escape poor local optima.

In addition, we use a differentiable soft Dice loss.
Let $P_c : \Omega \rightarrow [0,1]$ be the warped probabilistic logits for class $c$ after softmax, and $Y_c : \Omega \rightarrow \{0,1\}$ be the one-hot encoding of $S_f$.
The per-class soft Dice is
\begin{equation}
\mathrm{Dice}^{\text{soft}}_c(P, Y)
= \frac{2 \sum_{\mathbf{x}} Y_c(\mathbf{x}) P_c(\mathbf{x}) + \varepsilon}
{\sum_{\mathbf{x}} Y_c(\mathbf{x})^2
+ \sum_{\mathbf{x}} P_c(\mathbf{x})^2 + \varepsilon},
\end{equation}
and the multi-class average is
\begin{equation}
\mathrm{Dice}^{\text{soft}}(P, Y)
= \frac{1}{K} \sum_{c=1}^K \mathrm{Dice}^{\text{soft}}_c(P, Y).
\end{equation}
The corresponding loss used in Eq.~\ref{grpo} is
\begin{equation}
\mathcal{L}_{\text{Dice}} = 1 - \mathrm{Dice}^{\text{soft}}(P, Y).
\end{equation}

\subsection{Network Architectures and Gaussian Heads}
\label{sec:arch}

We adopt the official implementations of VoxelMorph-L, TransMorph, and NICE-Trans as backbones and attach a lightweight Gaussian policy head on their top-level encoder features.

\paragraph{VoxelMorph-L.}
VoxelMorph-L is a symmetric 3D U-Net with encoder channels
$[32, 64, 128, 256, 256]$ and decoder channels
$[256, 256, 128, 64, 32]$.
We take the last encoder feature (before the bottleneck skip connection) as $f_L$.

\paragraph{TransMorph.}
For TransMorph, we follow the official 3D large variant, including its encoder–decoder hierarchy and transformer blocks, and use the final encoder feature map as $f_L$.

\paragraph{NICE-Trans.}
In NICE-Trans, moving and fixed volumes are encoded independently into 128-channel features, concatenated into a 256-channel tensor, and then fed into the decoder.
We place the Gaussian policy head on this concatenated feature map.

Approximate latent dimensionalities $N$ for each dataset–backbone combination are summarized in Table~\ref{tab:latent_dim}.

\begin{table}[t]
\centering
\caption{Approximate latent dimensionality $N$ for each dataset/backbone. 
All values are computed as $N = H_L W_L D_L C_L$ under the standard input resolutions 
($160\times 192\times 224$ for OASIS/LiTS and $160\times 192\times 192$ for Abdomen MR$\leftarrow$CT).}
\label{tab:latent_dim}
\scalebox{0.8}{
\begin{tabular}{lccc}
\toprule
Dataset & VoxelMorph-L & TransMorph & NICE-Trans \\
\midrule
OASIS / LiTS      & $53{,}760$  & $ 161{,}280$ & $53{,}760$ \\
Abdomen MR$\leftarrow$CT    & $ 46{,}080$  & $ 138{,}240$ & $46{,}080$ \\
\bottomrule
\end{tabular}}
\end{table}

\subsection{Dataset Splits and Pair Construction}
\label{sec:splits}

We follow Learn2Reg~2021 protocols whenever possible and construct image pairs consistently across backbones. Volumes/scans are first partitioned into disjoint train, validation, and test pools; pair lists are then sampled only within the corresponding pool. Consequently, no test volume appears in warm-up, GRPO, or validation pairs.

\paragraph{OASIS.}
All volumes are preprocessed and resampled to $160 \times 192 \times 224$.
From 414 training volumes we form 400/100/20 pairs for warm-up, GRPO, and validation from the training pool only.
The 19 official validation pairs serve as our test set and are never used for training or hyperparameter tuning.

\paragraph{LiTS.}
LiTS provides 131 contrast-enhanced CT scans with liver and tumor annotations; we use the whole-liver labels only.
After preprocessing and resampling to $160 \times 192 \times 224$, we construct 400/100/20/40 pairs for warm-up, GRPO, validation, and test, ensuring that the held-out test pool is fully disjoint from the training and validation pools.

\paragraph{Abdomen MR--CT.}
This task contains 8 paired MR--CT scans from TCIA and 90 unpaired scans (50 CT from BCV and 40 MR from CHAOS), all resampled to $160 \times 192 \times 192$ with standard intensity preprocessing.
From the unpaired scans we form 400/100/20 MR--CT pairs for warm-up, GRPO, and validation, and use the 8 official paired scans as the test set.

For label-efficiency experiments, we subsample the 100 labeled training pairs at different sizes with fixed random seeds.
Unless otherwise stated, all refactored backbones share the same pair lists on each benchmark.

\subsection{Training Protocols and Hyperparameters}
\label{sec:train}

\paragraph{General optimization.}
Unless noted, all models are trained with Adam, learning rate $10^{-4}$, and batch size 1 3D pair.
The same learning-rate scale is used for warm-up and GRPO.

\paragraph{Fairness across baselines.}
VoxelMorph-L, TransMorph, and NICE-Trans are refactored under the same weakly supervised setting and trained on identical pair lists with the same labeled pairs and comparable epoch budgets. CorrMLP, RIIR, SPAC, and WarpDDF+RegCut are reproduced by following their released code or published protocols when a full unification is not directly supported; we report their results under those settings in the main paper.

\paragraph{Gaussian head constraints.}
The two $1 \times 1 \times 1$ convolutional heads that output $\boldsymbol{\mu}$ and $\log \boldsymbol{\sigma}$ are regularized as in Eqs.~\ref{mu}–\ref{logsigma} of the main paper.
A representative setting (TransMorph on OASIS) uses
$\lambda_{\text{scale}} = 10$, $\sigma_{\min} = -10$, $\sigma_{\max} = 3$.

\paragraph{Warm-up stage.}
Warm-up optimizes the loss in Sec.~\ref{sec:loss}.
For TransMorph on OASIS we use $\lambda_{\text{reg}} = 1$ and $\beta_{\text{KL}} = 10^{-4}$; other dataset–backbone combinations choose values of the same order.
We run warm-up such that each training pair is seen roughly ten times (about $O(50)$ epochs under our standard settings) and select the checkpoint with the best validation Dice for subsequent GRPO.

\paragraph{GRPO stage.}
GRPO uses the latent-space policy in Sec.~3.3 with the reward in Sec.~\ref{sec:reward}.
In our main configuration (e.g., TransMorph on OASIS), we use $J=\text{Trajs}=6$ trajectories per state and $T=\text{Steps}=3$ refinement steps.
Typical reward weights are $w_{\text{Dice}} = 10$ and $w_{\text{NJD}} = -100$.

The overall GRPO loss is
\begin{equation}
\mathcal{L}_{\text{GRPO}}
= \mathcal{L}_{\text{policy}}
+ \lambda_{\text{warm}} \mathcal{L}_{\text{warm}}
+ \lambda_{\text{Dice}} \mathcal{L}_{\text{Dice}},
\end{equation}
with $\lambda_{\text{warm}} = 0.8$ and $\lambda_{\text{Dice}} = 0.2$ in all main experiments.
The exploration temperature $\tau$ is linearly annealed from $\tau_{\text{init}} = 10$ to $\tau_{\min} = 2$ (e.g., decreasing by 1 every 10 epochs).
GRPO typically traverses the training set on the order of 30–60 epochs; final models are selected by the best validation Dice.

\subsection{Hardware and Software Environment}
\label{sec:env}

Experiments are conducted on a Linux cluster with up to eight NVIDIA A800-SXM4-80GB GPUs and dual Intel Xeon Silver 4316 CPUs per node.
We use Ubuntu~20.04.6 LTS, Python~3.x, and PyTorch~2.3.0.
Medical image I/O and preprocessing rely on SimpleITK~2.5.2 together with standard NumPy and SciPy utilities.

\subsection{Efficiency Analysis}
\label{sec:efficiency_supp}

Table~\ref{tab:efficiency_supp} reports the structural and runtime overhead introduced by the RL-friendly refactoring on OASIS. Across all three backbones, MorphSeek adds less than 3\% parameters, keeps single-step inference close to the original models, and exhibits near-linear latency growth with the number of refinement steps.

\begin{table}[t]
\centering
\caption{Efficiency analysis on OASIS. MorphSeek adds less than 3\% parameters and near-linear runtime growth with refinement steps.}
\label{tab:efficiency_supp}
\resizebox{0.9\linewidth}{!}{%
\begin{tabular}{l|rrr|rr}
\toprule
\multirow{2}{*}{\textbf{Baseline}} & \multicolumn{3}{c|}{\textbf{Model Parameters}} & \multicolumn{2}{c}{\textbf{GPU Inference Time (ms)}} \\
\cmidrule(lr){2-4} \cmidrule(lr){5-6}
& \textbf{Original} & \textbf{+$\Delta$ Abs} & \textbf{+$\Delta$ Rel} & \textbf{Original} & \textbf{+MorphSeek 1/2/3 step(s)} \\
\midrule
VoxelMorph-L & 27.05M & +0.13M & \cellcolor{green!15}+0.48\% & 625 & 685 / 1387 / 2022 \\
TransMorph & 46.77M & +1.18M & \cellcolor{green!15}+2.53\% & 401 & 444 / 900 / 1376 \\
NICE-Trans & 5.71M & +0.13M & \cellcolor{green!15}+2.27\% & 406 & 431 / 864 / 1295 \\
\bottomrule
\end{tabular}%
}
\end{table}

\begin{figure*}[t]
  \centering
  %\fbox{\rule{0pt}{2in} \rule{0.8\linewidth}{0pt}}
   \includegraphics[width=\linewidth]{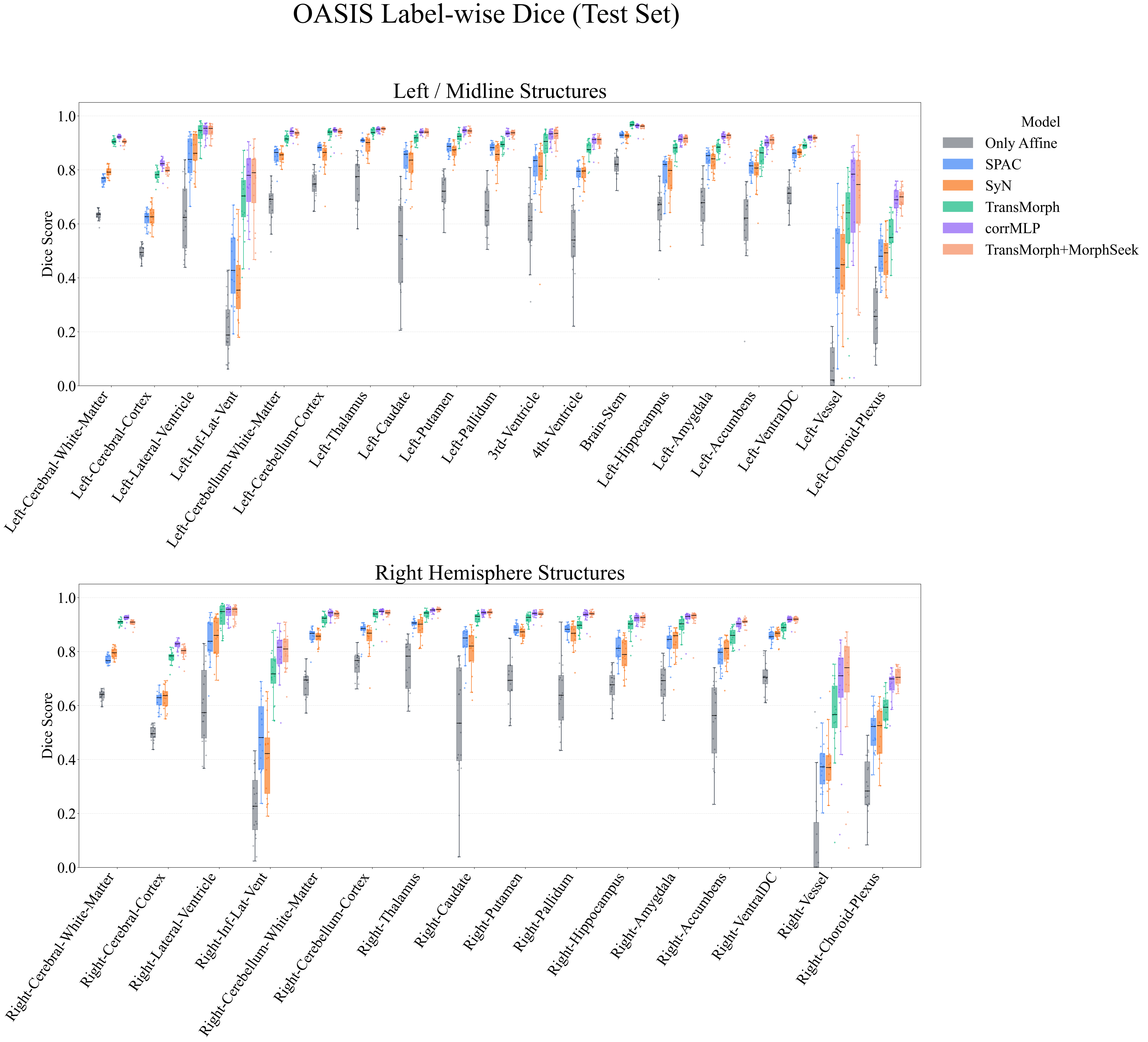}
   \caption{Label-wise Dice on OASIS (SPAC: Steps = 20, TransMorph+MorphSeek: Steps/Trajs = 3/6)}
   \label{10}
\end{figure*}

\begin{figure*}[t]
  \centering
  %\fbox{\rule{0pt}{2in} \rule{0.8\linewidth}{0pt}}
   \includegraphics[width=\linewidth]{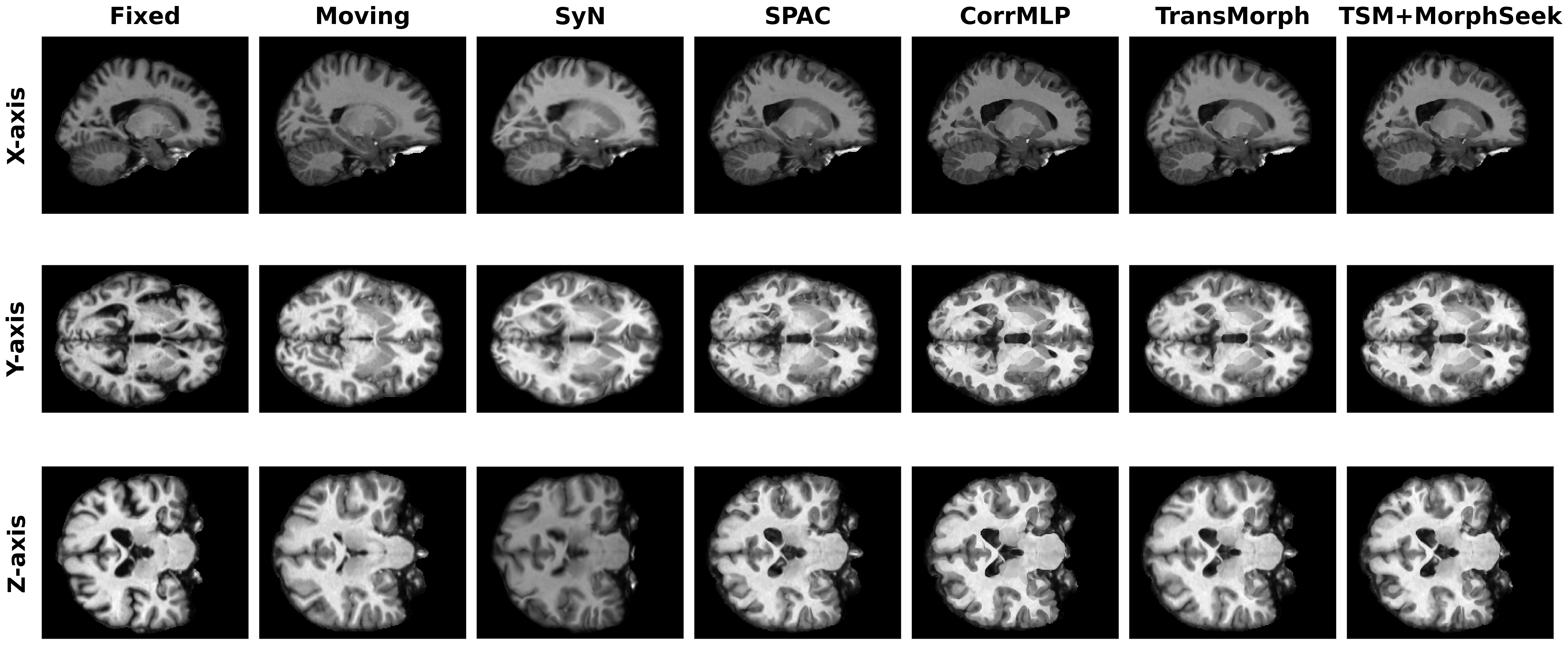}
   \caption{A Qualitative Registration Example on OASIS (SPAC: Steps = 20, TransMorph+MorphSeek: Steps/Trajs = 3/6)}
   \label{11}
\end{figure*}

\section{Additional Visual Results}
\label{sec:additional_visuals}

To complement the quantitative results in the main paper, we provide two additional visualizations on the OASIS brain MRI benchmark. Figure~\ref{10} reports label-wise Dice distributions on the test set for SyN, SPAC, CorrMLP, TransMorph, and TransMorph+MorphSeek. Figure~\ref{11} shows a representative registration example, comparing the fixed and moving images with the warped outputs of these methods in three orthogonal views.

\end{document}